\documentclass[11pt]{article}

\usepackage[a4paper, margin=1in]{geometry}
\usepackage{caption}
\date{}

\usepackage[utf8]{inputenc} %
\usepackage[T1]{fontenc}    %
\usepackage{hyperref}       %
\usepackage{url}            %
\usepackage{booktabs}       %
\usepackage{amsfonts}       %
\usepackage{nicefrac}       %
\usepackage{microtype}      %
\usepackage{xcolor}         %
\usepackage{graphicx} %
\usepackage{multirow}
\usepackage[export]{adjustbox}
\usepackage{amsmath}

\usepackage{etoolbox} %
\newbool{includeapp}
\setbool{includeapp}{true} %

\newcommand{\nside}{n_{\mathrm{side}}}

\usepackage[backend=biber,style=numeric,sorting=none]{biblatex}
\DeclareFieldFormat{urldate}{}

\DeclareSourcemap{
  \maps[datatype=bibtex]{
    \map{
      \step[fieldset=primaryclass, null]
    }
  }
}

\title{PEAR: Equal Area Weather Forecasting on the Sphere}

\author{
Hampus Linander\textsuperscript{ 1, 2}\;\; Tage Tykesson\textsuperscript{ 2, 4}\;\; Pietro Rosso\textsuperscript{ 3} \\
Christoffer Petersson\textsuperscript{ 2, 5}\;\; Daniel Persson\textsuperscript{ 2} \;\; Jan E.\ Gerken\textsuperscript{ 2} \\
\vspace{0.2cm} \\
\textsuperscript{1 }\begin{tabular}[t]{@{}l}
VERSES AI\\
Los Angeles, USA\\
\end{tabular}
\textsuperscript{2 }\begin{tabular}[t]{@{}l}
Department of Mathematical Sciences\\
 Chalmers University of Technology\\
 University of Gothenburg, Sweden \\
\end{tabular} \vspace{0.2cm} \\
\begin{tabular}[t]{@{}l}
\textsuperscript{3 }Department of Computer Science and Engineering\\
\textsuperscript{4 }Department of Physics and Astronomy \\
\phantom{\textsuperscript{4 }}Chalmers University of Technology, Sweden \\
\end{tabular}
\textsuperscript{5 }\begin{tabular}[t]{@{}l}
Recohere \\
Gothenburg, Sweden\\
\end{tabular} 
}

\begin{document}

\maketitle
\vspace{-3em}
\let\thefootnote\relax\footnotetext{Correspondence: \href{mailto:linander@chalmers.se}{linander@chalmers.se} \& \href{mailto:gerken@chalmers.se}{gerken@chalmers.se}}

\begin{abstract}
Artificial intelligence is rapidly reshaping the natural sciences, with weather forecasting emerging as a flagship AI4Science application where machine learning models can now rival and even surpass traditional numerical simulations. Following the success of the landmark models Pangu Weather and Graphcast, outperforming traditional numerical methods for global medium-range forecasting, many novel data-driven methods have emerged. A common limitation shared by many of these models is their reliance on an equiangular discretization of the sphere which suffers from a much finer grid at the poles than around the equator. In contrast, in the Hierarchical Equal Area iso-Latitude Pixelization (HEALPix) of the sphere, each pixel covers the same surface area, removing unphysical biases. Motivated by a growing support for this grid in meteorology and climate sciences, we propose to perform weather forecasting with deep learning models which natively operate on the HEALPix grid. To this end, we introduce Pangu Equal ARea (PEAR), a transformer-based weather forecasting model which operates directly on HEALPix-features and outperforms the corresponding model on an equiangular grid, and other baselines, without any computational overhead. Furthermore, we perform numerical experiments on the equivariance properties of our setup and verify the performance of PEAR on climate model emulation.\\
    Code: \url{https://github.com/hlinander/PEAR-Weather}.
\end{abstract}
\vspace{-2em}
\begin{figure}[h!]
    \centering
    \raisebox{-.5\height}{\includegraphics[width=0.3\linewidth]{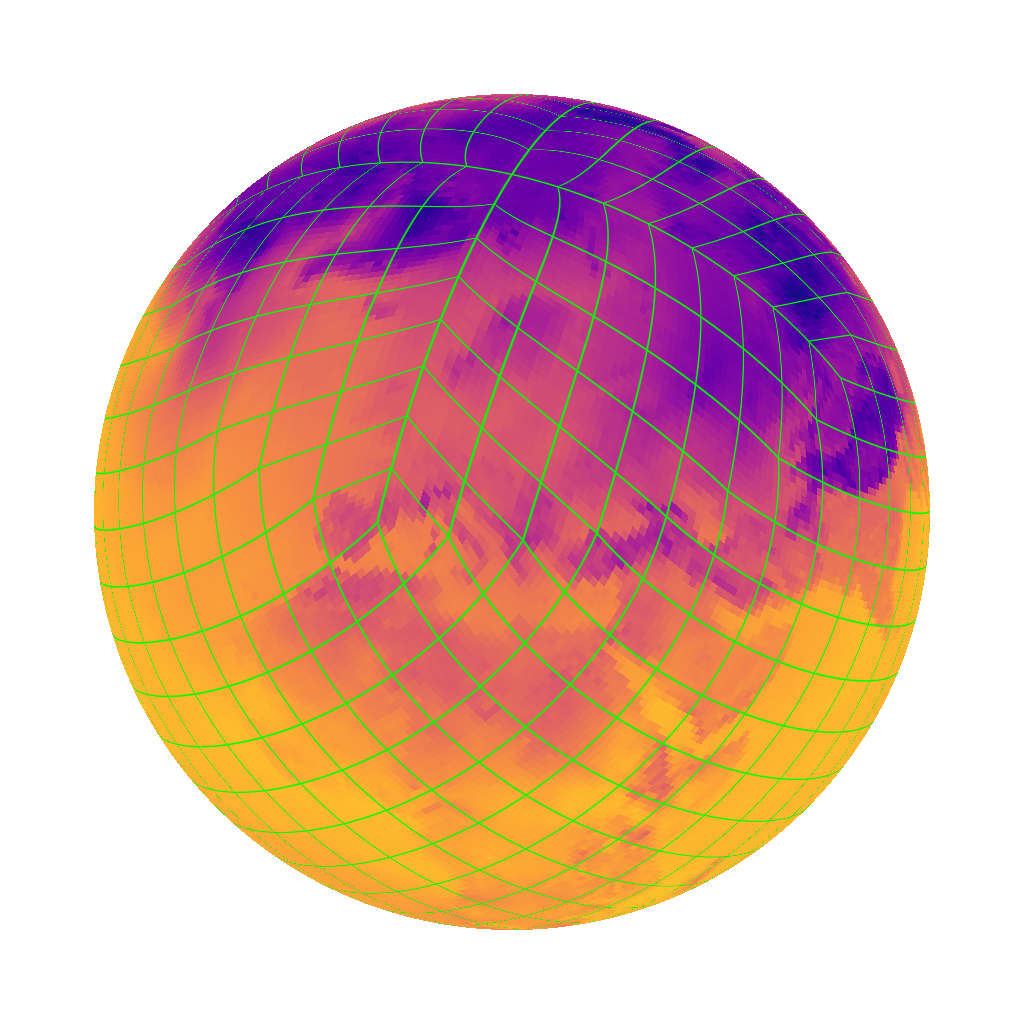}}
    \raisebox{-.5\height}{\includegraphics[width=0.4\linewidth]{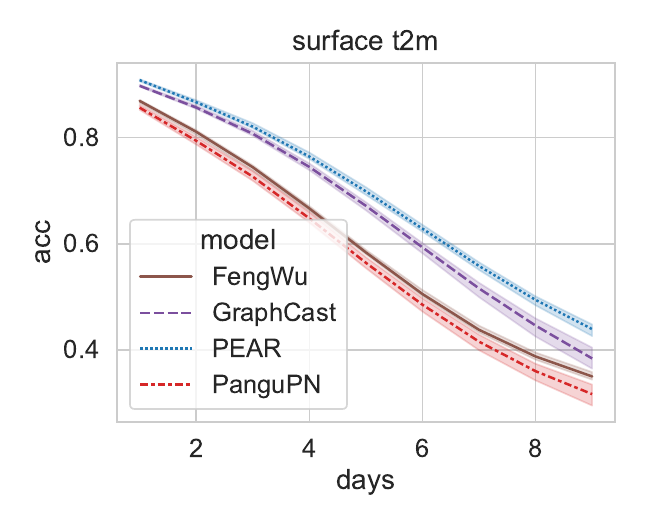}}
    \caption{Left: Predicted surface level temperature from PEAR. Green lines show the HEALPix cell boundaries at 3 levels of course-graining above the model resolution. Right: Anomaly correlation coefficient (ACC) (higher is better) for surface level temperature at 2m (t2m) with forecast horizon up to 10 days. PEAR outperforms all baselines.
    Shaded bands show the min-max range across 5 ensemble members.}
    \label{fig:prediction}
\end{figure}

\section{Introduction}
The convergence of artificial intelligence and the natural sciences—often referred to as AI4Science—is transforming how complex physical systems are studied and predicted. In Earth system science, machine learning approaches have achieved performance on par with or exceeding that of state-of-the-art numerical models~\cite{yu2024deep,liu2024foundation,zhao2024comprehensive}. Weather forecasting stands at the forefront of this shift~\cite{lam2023,chen2023,chen2023a,andrychowicz2023a,NEURIPS2024_7f19b99e}, driven by advances in architectures that can capture multiscale atmospheric dynamics directly from data, while offering orders of magnitude speedups compared to traditional numerical integration methods. Yet, most existing AI-based forecasting systems inherit discretizations from the numerical models they aim to complement or replace, such as the equiangular latitude-longitude and Gaussian grids, which imposes non-uniform resolution and unphysical biases. In this paper we explore the Hierarchical Equal Area iso-Latitude Pixelization (HEALPix) as a native grid for learning-based weather forecasting, equal area modeling across the globe.

Following the publication of the landmark models Pangu-Weather~\cite{bi2023} and Graphcast~\cite{lam2023}, the recent emergence of performant data-driven models for weather forecasting has been driven by the widespread use of the ERA5 reanalysis dataset~\cite{hersbach2020era5}, a comprehensive open access global climate data record produced by the European Center for Medium-Range Weather Forecasts (ECMWF).

ERA5 provides hourly estimates of atmospheric, land, and oceanic variables on a global scale, discretized using the regular equiangular latitude-longitude grid or a reduced Gaussian grid. Both of these grids introduce challenges when used in physics-based ML models. Specifically, both grids exhibit non-uniform resolution across latitudes, which can lead to artifacts and unphysical inductive biases in learning-based models.

To address these limitations, alternative spherical representations have been explored. One notable example is HEALPix, originally developed in astrophysics for uniform pixelization of the celestial sphere~ \cite{gorski1998analysis}. HEALPix offers equal-area pixel distributions and hierarchical resolution capabilities, making it well-suited for learning tasks on spherical domains. Recent efforts have begun to investigate the use of HEALPix for resampling next-generation weather and climate data~\cite{karlbauer2024,ramavajjala2024}.  HEALPix has also been used in other machine learning contexts where spherical data is natural~\cite{Perraudin_2019,carlsson2023}.

The ECMWF are also targetting the HEALPix grid in its Destination Earth (DestinE) initiative~\cite{egusphere-2025-2198,HOFFMANN2023100394}, which aims to develop a digital twin of the Earth for improved climate and weather predictions. By employing HEALPix, DestinE seeks to achieve high-resolution, globally consistent datasets that can better inform decision-making processes related to climate change adaptation and disaster risk management. Furthermore, ECMWF's transition from GRIB1 to GRIB2 data formats supports the integration of advanced grid systems like HEALPix~\cite{grib1togrib2}, enabling more detailed and efficient data representation essential for next-generation forecasting models.

Motivated by these developments, we propose PEAR (Pangu Equal Area), the first machine learning weather forecasting model that operates entirely on the HEALPix grid. Unlike previous approaches that project onto planar grids or only use HEALPix for preprocessing, PEAR natively represents inputs, internal features, and outputs on the spherical HEALPix domain, enabling consistent resolution across the globe and better alignment with the underlying physical symmetries. We train on the ERA5-Lite  dataset~\cite{bi2023}, a subset of the full ERA5 data, resampled to HEALPix, as the forecasting target and evaluate our model's performance on key atmospheric variables across various resolutions.

\newpage
Our main contributions are:
\begin{itemize}
    \item Motivated by an increased adoption of the HEALPix grid in next generation weather and climate data, we propose to use the spherical HEALPix grid as the native grid for machine learning weather predictions. This approach eliminates unphysical biases in the sampling of the sphere affecting  standard equiangular based weather prediction models and removes the need for spatial weights in the loss and evaluation metrics.
    \item We introduce PEAR: a baseline model for neural weather simulation on HEALPix using a volumetric transformer architecture which operates at no computational overhead compared to an equivalent model on the traditional equiangular grid.
    To demonstrate the advantage of using a native HEALPix model, we show that PEAR's HEALPix-predictions outperform those produced by the same architecture operating on an equiangular grid, as well as other baseline models.
    \item We evaluate all models for lead times of up to ten days and show that this advantage persists for longer forecasting horizons. We verify that daily sampling in the ERA5-Lite dataset does not alter the equivariance properties of the weather prediction task. Furthermore, we demonstrate that PEAR shows competitive performance on climate model emulation for a large number of climate models.
\end{itemize}

\section{Related work}

The field of machine-learning weather forecasting has received tremendous attention over the last years both for medium-range weather forecasting~\cite{shu2024} and extreme weather prediction~\cite{olivetti2024}. The first model whose performance on global medium-range forecasts surpassed that of numerical models was Pangu-Weather~\cite{bi2023} which is based on a volumetric version of the SWIN transformer~\cite{liu2021a}. Since then, a number of models have been published which improved upon this baseline, such as GraphCast~\cite{lam2023}, FuXi~\cite{chen2023}, FengWu~\cite{chen2023a}, NetMet-3~\cite{andrychowicz2023a}, Stormer~\cite{NEURIPS2024_7f19b99e} or the ECMWF's data-driven forecasting system~\cite{lang2024}. Fourier neural operators have been used in a number of models in this domain~\cite{kurth2023, bonev2023} as well. In order to take the curvature of the sphere into account, one model~\cite{PhysicsInspiredAdaptionstoLowParameterNeuralNetworkWeatherForecastSystems} used Spherenet. Similarly, the forecasting model CirT~\cite{liu2025cirt} is based on a circular transformer which takes the azimuthal circularity of the sphere into account. Probabilistic weather forecasting models allow for uncertainty estimation of the generated predictions~\cite{oskarsson2024}. The model GenCast creates an ensemble of stochastic predictions which outperforms the top operational medium-range weather forecast in the world, ENS, the ensemble forecast of the European Centre for Medium-Range Weather Forecasts~\cite{price2024}. The consideration of physical conservation laws in the training process can improve these data-driven weather prediction models~\cite{sha2025}. Closely related to machine-learning weather forecasting systems are neural network models which predict the climate~\cite{nguyen2023a, watt-meyer2023} or general-purpose models for the earth system~\cite{bodnar2024}. Contributions to the training setup include careful ablations of various aspects of the architecture~\cite{AnalyzingandExploringTrainingRecipesforLargeScaleTransformerBasedWeatherPrediction} and a training platform for deep-learning based weather prediction models~\cite{schreck2024}. WeatherBench 2~\cite{rasp2024} provides a well-established benchmark for machine-learning weather prediction models. Deep neural networks have also been applied in the post-processing of global weather forecasts~\cite{bouallegue2024}. A hybrid model of machine-learning components and a differentiable solver for circulation models was proposed in~\cite{kochkov2024}.

The aforementioned models have in common that they use the naive equiangular discretization of the sphere as input. The resulting grid is considerably denser towards the poles than around the equator. In contrast, the HEALPix grid is uniform over the entire sphere. The convolutional DLWP-HPX model~\cite{karlbauer2024} uses representations on the HEALPix grid. HEAL-ViT~\cite{ramavajjala2024} is a vision-transformer based model which operates on the HEALPix grid and uses a learnable encoder and decoder to map the features from the equiangular grid to HEALPix and back. In contrast, our model operates entirely on HEALPix, in line with our proposal to use the HEALPix grid as the physically appropriate grid for weather forecasting.

\section{HEALPix}
The HEALPix grid uniformly divides the sphere into four-sided polygons (quadrilaterals) of equal area. The pixels are positioned at the centers of the quadrilaterals and lie on circles of constant latitude with equal spacing in azimuthal angle. The construction of the grid starts from 12 base-quadrilaterals, 4 equally shaped quadrilaterals grouped around each pole and 4 equally shaped quadrilaterals around the equator. These base-quadrilaterals are then repeatedly divided into two equally-sized halves along their edges. After $k$ divisions, there are $\nside=2^k$ pixels along each edge of each base quadrilateral, resulting in a grid with $12\cdot\nside^2$ pixels in a hierarchical structure.

For computations, the pixels are organized in a one-dimensional list. In the \emph{nested} ordering, the pixel indices are provided by the hierarchical construction of the grid such that merging blocks of four consecutive pixels in the list coarse-grains the grid from $\nside$ to $\nside/2$. We will use this property for coarse- and fine-graining the pixel grid. Similarly, blocks of $4^k$ consecutive pixels in the nested ordering correspond to the pixels in quadrilaterals $k$ division-levels above the grid resolution. We will use this property to easily divide the surface of the sphere into attention-windows.

In the \emph{ring} ordering of the pixel list, the pixels are sorted along the iso-latitude circles, from the north pole to the south pole. Performing a roll operation on this list rotates the features on the sphere around the polar axis. Around the poles, the features will additionally be distorted due to the decreasing number of pixels per iso-latitude circle. We will use this roll operation in the ring ordering to shift the features between windows after attention layers. For a roll by $n$ pixels, the last $n$ pixels around the south pole will spill over to the north pole and therefore be masked in the attention weights.

For conversions between the ring- and nested indexing of the pixels and to retrieve the pixel positions in spherical coordinates, we use the Python bindings for the HEALPix package provided by \texttt{chealpix}.

\subsection{ERA5 on HEALPix}
The ERA5 dataset~\cite{hersbach2020era5} contains the global atmospheric state in terms of a number of
hydrodynamical quantities, discretized on an equiangular gridding on the
sphere, over multiple vertical slices and hourly in time. Following Pangu-Weather~\cite{bi2023},
we distinguish between surface variables (temperature, wind velocity, pressure,
humidity) and upper variables (temperature, wind velocity, humidity, pressure).

The raw data provided by ECWMF has an angular gridding of $0.25^\circ$ resulting in a spatial
$(\mathrm{lon}, \mathrm{lat})$ resolution of $(1440, 721)$. Since the
equiangular grid cells are not of equal area, the effective spatial
resolution on the sphere is not constant. The lowest spatial resolution can be
found on the equator, where each cell has an angular area of $1.9\cdot 10^{-5} \mathrm{rad}^2$. We target a HEALPix grid with $\nside = 64$, where all of the $12 \cdot 64^2$ pixels cover an
angular area of $2.6\cdot 10^{-4}$, so that the ERA5 data has a higher spatial
resolution everywhere on the globe and saturates the HEALPix pixel density.

\subsection{ERA5-lite}
\label{sec:era5lite}
The full ERA5 dataset contains 40 years with hourly intervals, resulting in multiple petabytes of storage required. To facilitate research on weather models, a reduced dataset using 11 years and 24 hour intervals has been used~\cite{bi2023}. This subset consists of the years 2007 to and including 2017, with 2019 used for validation. All samples are at 00:00UTC, resulting in a total of 4017 training samples and 365 validation samples. Together with cached normalized data, the total dataset size is about 3TB.

\section{PEAR: Pangu Equal Area}
\label{sec:pear}
The weather forecasting task is formulated as a  regression problem, where the input is the global weather state at time $t$ and the output is the global weather state at time $t + \Delta t$. Here we use a time delta of 24 hours.

The global volumetric weather state is discretized on the HEALPix grid along the surface, and into 13 discrete levels in the vertical direction.
Following prior work~\cite{bi2023}, we represent the total weather state as a combination of 4 surface variables (wind speed along the surface, temperature and mean sea level pressure), and 5 upper variables (wind speed along the sphere, temperature, specific humidity and geopotential) at 13 discrete vertical levels. PEAR thus takes two input tensors, the surface and upper variables discretized on the spherical surface and the spherical shell correspondingly.
Since the HEALPix grid covers the sphere with a 1d index structure, the model input tensors have shape $(12 n_{\mathrm{side}}^2, 4)$ and $(12 n_{\mathrm{side}}^2, 13, 5)$.

The architecture is constructed using a combination of 5 main layer types: patch embedding, windowed attention with alternating shifting, downsampling, upsampling and patch recovery. See Figure \ref{fig:arch} for a schematic overview of the architecture, and Table \ref{tab:arch} in the appendix for details on the layer parameters.

\subsection{Patch embedding}
The initial patch embedding uses a 1d convolution with kernel size 16 and stride 16 for the surface variables, and a 2d convolution with kernel size and stride $(16, 2)$ for the upper variables. This corresponds to a patch size of $4\times4$ in an equiangular grid. Both convolutions output 48 channels. The patch embedded tensors have shape $(\nicefrac{3}{4} n_{\mathrm{nside}}^2, 48)$ and $(\nicefrac{3}{4} n_{\mathrm{nside}}^2, 7, 48)$. At this point the patch embedded surface variables are concatenated to the patch embedded upper variables, resulting in a single tensor of shape $(\nicefrac{3}{4} n_{\mathrm{nside}}^2, 8, 48)$.

\subsection{Windowed attention}
We partition the tensors into windows using the nested structure of the HEALPix grid. The vertical direction is also partitioned into neighbouring levels. In terms of the latent tensor shape $(\nicefrac{3}{4} n_{\mathrm{nside}}^2, 8, 48)$, partitioning using a window size of $(W_{\mathrm{hp}}, W_d)$ gives a windowed tensor of shape $(\frac{\nicefrac{3}{4} n_{\mathrm{nside}}^2}{W_{\mathrm{hp}}} \frac{8}{W_\mathrm{d}}, W_{\mathrm{hp}} W_d, 48)$. Attention is now performed over the embeddings of the $W_{\mathrm{hp}} W_d$ voxels in each window. Note that the nested index structure of the HEALPix grid naturally supports the window partitioning in terms of contiguous memory. See Figure \ref{fig:patch_and_window} for an illustration of the patches and first level windows corresponding to the tensor structure at the first attention layer.

In contrast to Pangu~\cite{bi2023}, we use a simplified learned relative positional embedding. Because of the equal area grid cells in the HEALPix grid we can share relative positional embeddings between windows, and thus simply learn a relative positional embedding tensor $B$ of shape $(1, N_{\mathrm{heads}}, \left(W_d W_{\mathrm{hp}}\right)^2)$. The final attention computation is thus
$\mathrm{Att}(Q, K, V) = \mathrm{SoftMax}\left(\frac{Q K^\top}{\sqrt{d}} + B\right) V$,
where $d$ is the embedding dimension. The simplified positional embedding accounts for most of the parameter savings compared to Pangu in Table~\ref{tab:model_size}.

\begin{figure}
    \centering
    \includegraphics[width=0.5\textwidth]{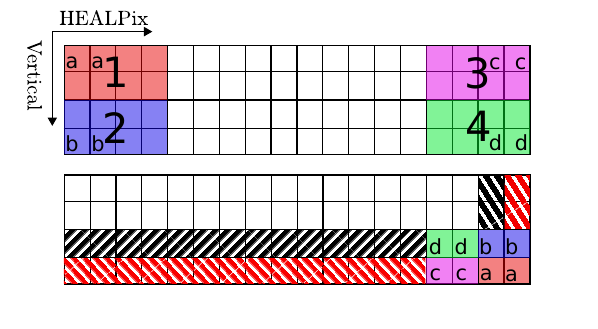}
    \caption{Shift and corresponding mask for windowed attention. Illustration of a scalar tensor with the 1d HEALPix index in the horizontal direction and the vertical direction corresponding to the discretized vertical direction above the surface. Ring indexing is used in the HEALPix direction, with the north pole to the left, and south pole to the right. The lower grid shows where the indicated voxels from the top grid end up after a negative shift of half a window in both directions. Note that the voxels from region a, b, c and d are all from spatially separated regions, and thus the window in the lower right needs to be masked accordingly. The colored striped regions indicate the mask for windows along the borders, where each window contains two regions instead of four as in the corner window.\label{fig:shift_mask}}
\end{figure}

To propagate information between the windows, that are otherwise disjoint in terms of attention, we shift the grid by roughly half the window size every other attention layer. Along the spherical directions of the HEALPix grid we employ the ring shifting strategy of HEAL-SWIN~\cite{carlsson2023}, and a simple shift in the vertical direction. Since this shifting is performed cyclically, there will be voxels in the polar regions that will jump from the north to the south pole, and from the lowest upper level to the highest upper level. To prevent attention among these spatially disjoint voxels, we implement masked attention. See Figure \ref{fig:shift_mask} for an illustration of the masking pattern that arises from the above shifting strategy. Note that because of the 1d structure of the HEALPix grid, this figure is representative for the actual tensor structure used in the implementation. To facilitate the shifting in the ring indexing scheme, we precompute the index conversion from nested to ring scheme.
Each windowed attention block follows the structure of SWIN-V2~\cite{liu2021swinv2} using layer-norm and skip connection.

\subsection{Downsampling and upsampling}
To facilitate a bottleneck structure, we follow Pangu~\cite{bi2023} and perform a single downsampling along the spherical directions. The HEALPix grid has a hierarchical structure where four neighbouring grid cells combine into a single grid cell at a coarser resolution. This provides a natural downsampling by concatenating the embeddings of groups of four neighbouring pixels, and then linearly projecting to the target embedding dimension. This downsampling is efficient in the nested indexing of the HEALPix grid, where this simply corresponds to a reshaping of the tensor, followed by a linear layer for projection.

The upsampling layer follows the same logic in reverse: we first expand the embedding dimension of a voxel to four times the target embedding dimension, followed by a reshaping into four new voxels along the HEALPix grid in the nested indexing scheme.

\subsection{Patch recovery}
To recover the surface and upper variable tensors, we use  transpose convolutions on the first and remaining vertical levels correspondingly. The latent tensor $x$ of shape $(\nicefrac{3}{4} n_{\mathrm{side}}^2, 8, 48)$ is split along the second dimension into a surface latent $x_{\mathrm{surface}}$ of shape $(\nicefrac{3}{4} n_{\mathrm{side}}^2, 1, 48)$, and an upper latent tensor $x_{\mathrm{upper}}$ of shape $(\nicefrac{3}{4} n_{\mathrm{side}}^2, 7, 48)$. To recover the output surface variables we perform a 1d transpose convolution on $x_{\mathrm{surface}}$ with channel count 4. The output upper variables are recovered with a 2d transpose convolution with channel count 5.
\begin{figure}[t]
    \centering
    \includegraphics[width=0.95\linewidth]{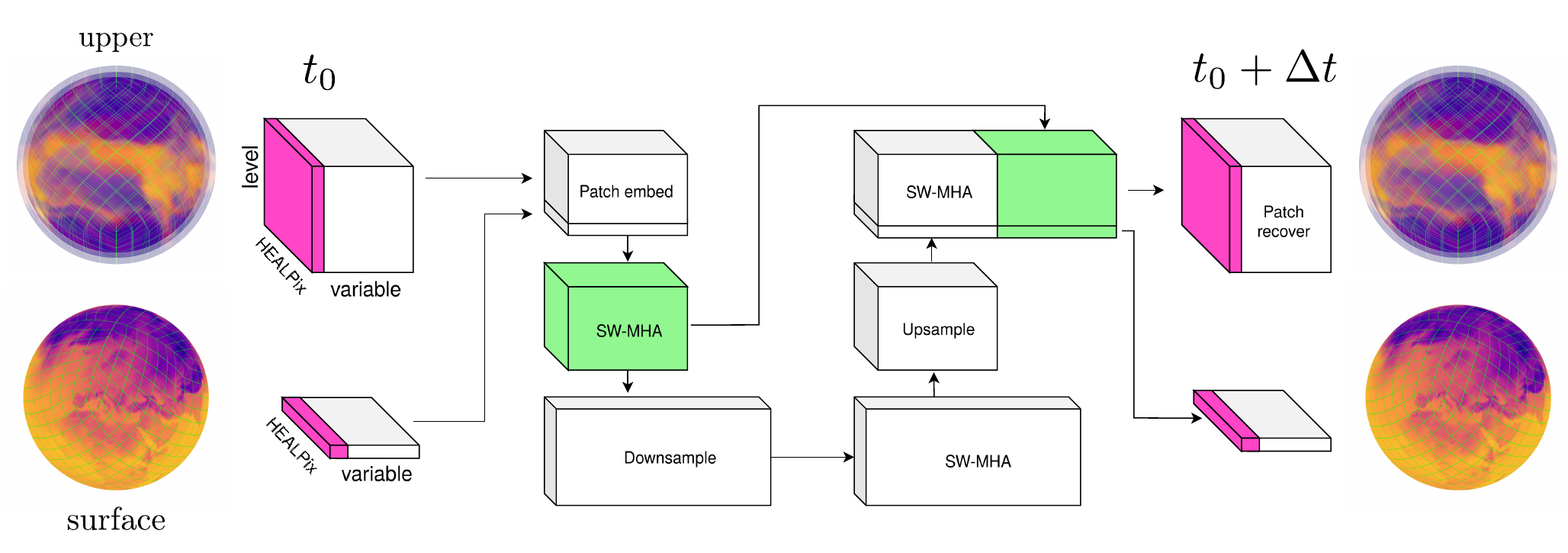}
    \caption{PEAR architecture schematic. Violet slices correspond to the variables visualized on the sphere for the input and output tensors. Each block indicates the tensor shape after the layer with the corresponding name. Patch embedding by convolution, shifted windowed multi-head attention (SW-MHA) with learned positional embedding, downsampling and upsampling by patch merging and splitting and patch recovery by transpose convolutions. Green block indicates the skip connection, where the output of the first attention layer is concatenated along the embedding dimension before the final patch recovery by transpose convolutions. Green lines on the spherical visualizations indicate the HEALPix grid at 3 levels of course-graining above the model resolution.}
    \label{fig:arch}
\end{figure}

\section{Experiments}
We evaluate PEAR against Pangu~\cite{zhuoqun2025}, GraphCast~\cite{lam2022} and FengWu~\cite{chen2023a} using the ERA5-lite subset described in Section \ref{sec:era5lite}. All experiments are carried out using $n_{\mathrm{side}}=64$ to saturate the angular resolution of the HEALPix grid using the available ERA5 data on the equiangular grid. We train all models using L1 loss, with the loss contribution from the surface variables weighted by $\frac{1}{4}$. We use the AdamW optimizer with weight decay $3\times10^{-6}$ and learning rate $5\times 10^{-4}$. On a single A100, PEAR and Pangu converges in 15 hours, FengWu in 35 hours and GraphCast in 40 hours.

To evaluate the medium-term forecasting ability of PEAR, we perform iterated model inference up to 10 times, resulting in forward time predictions of up to 10 days. At each lead time we calculate the average RMSE and anomaly correlation coefficient (ACC)~\cite{van2017probability} of all variables over the globe. See Appendix \ref{appendix} for details.

We evaluate baseline-predictions on an equiangular grid by applying the latitude weighting used in prior work~\cite{bi2023}. The equal area grid cells of HEALPix make this reweighting redundant for PEAR. The ACC measures the correlation between deviations from the climatology mean of predicted and ground truth forecasts, with a value of 1 indicating perfect agreement~\cite{hersbach2020era5}.
The climatology average is subtracted to factor out seasonal variations that would otherwise improve the raw correlation between predictions and ground truth data.
To calculate the climatology mean, we average each day of the year over the 11 years in the ERA5-lite dataset for every variable.

Pangu provides a baseline against the same type of architecture where an equiangular grid is used for input, latent and output tensors. Since the overall architectural blocks are the same, we can match the depth, number of attention heads and embedding dimensions for each block.
To compare against architecturally different baselines, we also evaluate GraphCast~\cite{lam2023} and FengWu~\cite{chen2023a} using the implementations in the PhysicsNemo framework~\cite{physicsnemo2023}, trained on the same ERA5-lite data with identical loss and optimizer settings. GraphCast is parameter matched to PEAR. All models are trained with 5 seeds. For Pangu, GraphCast and FengWu we use an equiangular grid with $n_\mathrm{lon}=314$ and $n_\mathrm{lat}=157$, resulting in a relative difference of the total number of pixels of $0.3\%$, or 146 pixels.
\begin{table}[!ht]
    \centering
    \small
    \caption{Model size and inference times. Inference times measure a forward pass through the models with input on the GPU. Mean and standard deviation over 100 iterations after warm-up.}
    \label{tab:model_size}
    \begin{tabular}{lll}
        \toprule
         Model & Trainable parameters (M) & Inference time (ms) \\
         \midrule
         PEAR & $4.3$ & $17 \pm 0.12$\\
         Pangu & $11.3$ & $25 \pm 0.3$ \\
         GraphCast & $4.3$ & $27.32 \pm 0.05$ \\
         FengWu & $14.3$ & $98.85 \pm 1.45$ \\
         \bottomrule
    \end{tabular}
\end{table}

\begin{figure}[h]
    \centering
    \includegraphics[width=0.95\linewidth]{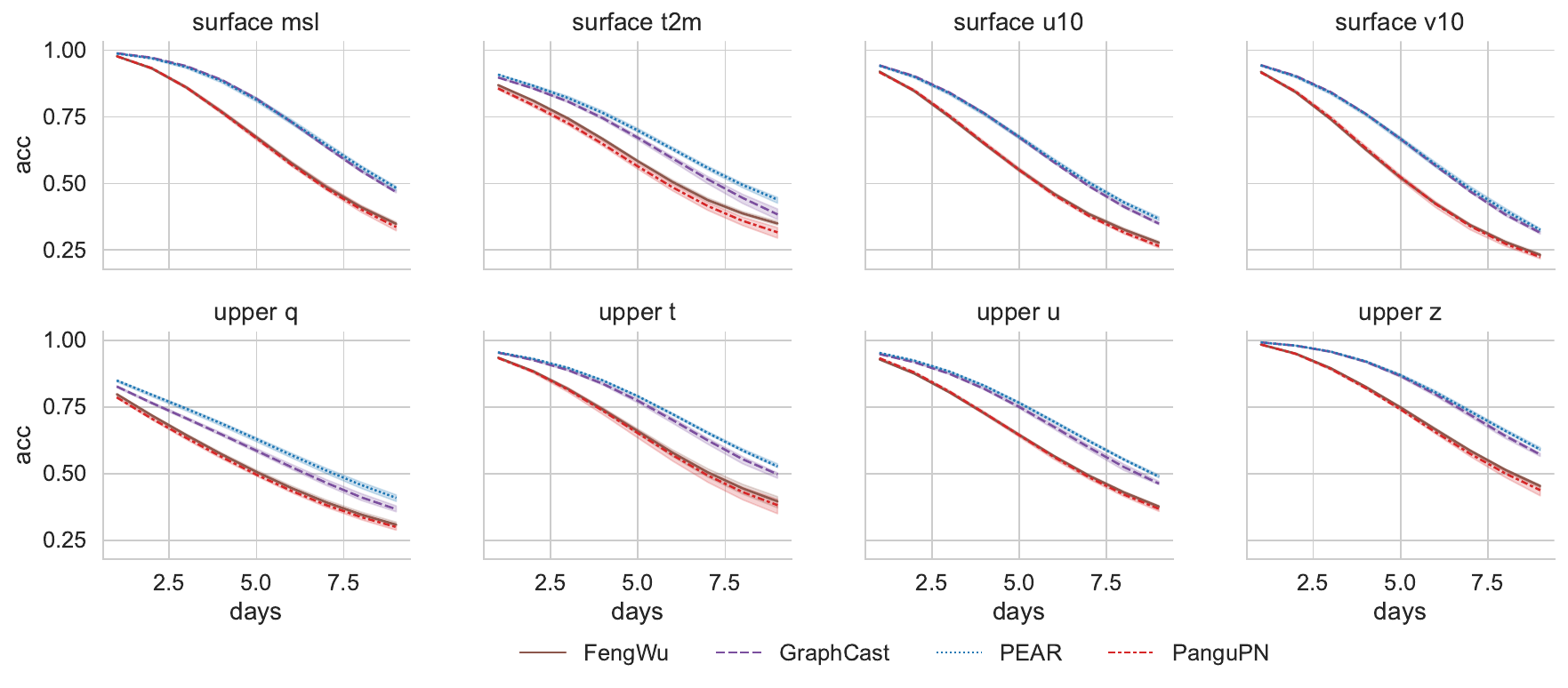}
    \caption{Mean anomaly correlation coefficient (ACC), higher is better, for the surface and upper variables after iterated model inference to perform multiple day forecasting. The upper variables are averaged over the 13 vertical levels. Shaded bands show the min-max range across 5 ensemble members. The metrics are mean sea level pressure (msl), temperature at 2m (t2m), eastward wind velocity at 10m (u10), northward wind velocity at 10m (v10), specific humidity (q), temperature (t), eastward wind velocity (u), northward wind velocity (v), geopotential (z). PEAR consistently outperforms all baselines across most variables and lead times.}
    \label{fig:acc_iterated}
\end{figure}

Figure \ref{fig:acc_iterated} shows the ACC for all surface and upper variables, with the ACC for the upper variables averaged over the 13 vertical levels. Shaded bands indicate the min-max range across 5 ensemble members. PEAR shows superior forecasting ability compared to all baselines across most variables and lead times. Table~\ref{tab:model_size} lists model sizes and inference times.

Table \ref{tab:acc} lists the average ACC for prediction lead times of 1, 5 and 9 days, see Table \ref{tab:rmse} in the appendix for the corresponding RMSE values. PEAR achieves higher ACC than Pangu and FengWu for every individual ensemble member across all variables and lead times. It also outperforms GraphCast for most variables, and in particular for longer prediction lead times, using only 63\% of the total inference time. Additional evaluations and raw model outputs are provided in the appendix.

\begin{table}[!ht]
\centering
\caption{Average anomaly correlation coefficient (ACC) for three different prediction lead times (1, 5 and 9 days). Higher value is better, best model mean in bold face. Range in sub- and superscript are min and max values over 5 model seeds. Models where the min/max range overlaps the best model are also shown in bold.}
\label{tab:acc}
\small
\renewcommand{\arraystretch}{1.4}
\begin{tabular}{l r r r r r l}
\toprule
Variable & $\Delta t$ & \multicolumn{4}{c}{ACC} & unit \\
\cmidrule(lr){3-6}
 & (days) & FengWu & GraphCast & PEAR & PanguPN & \\
\midrule
$_\mathrm{surface}^\mathrm{msl}$ & 1 & $0.977^{+0.001}_{-0.001}$ & $\mathbf{0.989^{+0.000}_{-0.000}}$ & $0.988^{+0.001}_{-0.001}$ & $0.977^{+0.001}_{-0.001}$ & \multirow{3}{*}{$\mathrm{Pa}$} \\
 & 5 & $0.673^{+0.006}_{-0.005}$ & $\mathbf{0.818^{+0.004}_{-0.003}}$ & $\mathbf{0.814^{+0.007}_{-0.008}}$ & $0.669^{+0.009}_{-0.007}$ &  \\
 & 9 & $0.348^{+0.007}_{-0.005}$ & $0.467^{+0.005}_{-0.003}$ & $\mathbf{0.483^{+0.006}_{-0.009}}$ & $0.336^{+0.013}_{-0.014}$ &  \\
\midrule
$_\mathrm{surface}^\mathrm{t2m}$ & 1 & $0.869^{+0.002}_{-0.004}$ & $0.898^{+0.002}_{-0.001}$ & $\mathbf{0.908^{+0.002}_{-0.002}}$ & $0.856^{+0.005}_{-0.004}$ & \multirow{3}{*}{$\mathrm{K}$} \\
 & 5 & $0.584^{+0.005}_{-0.004}$ & $0.671^{+0.007}_{-0.008}$ & $\mathbf{0.699^{+0.008}_{-0.007}}$ & $0.564^{+0.011}_{-0.010}$ &  \\
 & 9 & $0.350^{+0.008}_{-0.006}$ & $0.384^{+0.020}_{-0.019}$ & $\mathbf{0.439^{+0.009}_{-0.013}}$ & $0.316^{+0.018}_{-0.022}$ &  \\
\midrule
$_\mathrm{surface}^\mathrm{u10}$ & 1 & $0.920^{+0.002}_{-0.003}$ & $\mathbf{0.943^{+0.002}_{-0.001}}$ & $\mathbf{0.942^{+0.001}_{-0.003}}$ & $0.917^{+0.003}_{-0.003}$ & \multirow{3}{*}{$\frac{\mathrm{m}}{\mathrm{s}}$} \\
 & 5 & $0.551^{+0.005}_{-0.006}$ & $\mathbf{0.673^{+0.004}_{-0.005}}$ & $\mathbf{0.675^{+0.006}_{-0.008}}$ & $0.551^{+0.008}_{-0.006}$ &  \\
 & 9 & $0.278^{+0.005}_{-0.004}$ & $0.349^{+0.005}_{-0.004}$ & $\mathbf{0.367^{+0.008}_{-0.007}}$ & $0.265^{+0.006}_{-0.007}$ &  \\
\midrule
$_\mathrm{surface}^\mathrm{v10}$ & 1 & $0.919^{+0.002}_{-0.003}$ & $\mathbf{0.944^{+0.002}_{-0.001}}$ & $\mathbf{0.943^{+0.001}_{-0.003}}$ & $0.917^{+0.003}_{-0.002}$ & \multirow{3}{*}{$\frac{\mathrm{m}}{\mathrm{s}}$} \\
 & 5 & $0.522^{+0.005}_{-0.005}$ & $\mathbf{0.667^{+0.004}_{-0.002}}$ & $\mathbf{0.668^{+0.007}_{-0.010}}$ & $0.524^{+0.008}_{-0.011}$ &  \\
 & 9 & $0.232^{+0.005}_{-0.003}$ & $\mathbf{0.315^{+0.004}_{-0.005}}$ & $\mathbf{0.326^{+0.005}_{-0.007}}$ & $0.224^{+0.008}_{-0.006}$ &  \\
\midrule
$_\mathrm{upper}^\mathrm{q}$ & 1 & $0.798^{+0.003}_{-0.003}$ & $0.827^{+0.003}_{-0.004}$ & $\mathbf{0.850^{+0.002}_{-0.005}}$ & $0.787^{+0.004}_{-0.003}$ & \multirow{3}{*}{$\frac{\mathrm{g}}{\mathrm{kg}}$} \\
 & 5 & $0.507^{+0.007}_{-0.004}$ & $0.586^{+0.008}_{-0.007}$ & $\mathbf{0.630^{+0.006}_{-0.010}}$ & $0.496^{+0.011}_{-0.006}$ &  \\
 & 9 & $0.309^{+0.010}_{-0.006}$ & $0.367^{+0.012}_{-0.010}$ & $\mathbf{0.409^{+0.012}_{-0.013}}$ & $0.300^{+0.010}_{-0.012}$ &  \\
\midrule
$_\mathrm{upper}^\mathrm{t}$ & 1 & $0.936^{+0.002}_{-0.002}$ & $\mathbf{0.954^{+0.002}_{-0.001}}$ & $\mathbf{0.955^{+0.001}_{-0.001}}$ & $0.934^{+0.002}_{-0.002}$ & \multirow{3}{*}{$\mathrm{K}$} \\
 & 5 & $0.659^{+0.009}_{-0.010}$ & $0.774^{+0.007}_{-0.008}$ & $\mathbf{0.791^{+0.002}_{-0.004}}$ & $0.652^{+0.011}_{-0.018}$ &  \\
 & 9 & $0.396^{+0.018}_{-0.024}$ & $0.498^{+0.005}_{-0.015}$ & $\mathbf{0.527^{+0.010}_{-0.004}}$ & $0.382^{+0.022}_{-0.032}$ &  \\
\midrule
$_\mathrm{upper}^\mathrm{u}$ & 1 & $0.929^{+0.001}_{-0.002}$ & $0.949^{+0.002}_{-0.001}$ & $\mathbf{0.954^{+0.001}_{-0.002}}$ & $0.934^{+0.002}_{-0.002}$ & \multirow{3}{*}{$\frac{\mathrm{m}}{\mathrm{s}}$} \\
 & 5 & $0.646^{+0.004}_{-0.004}$ & $0.751^{+0.006}_{-0.006}$ & $\mathbf{0.766^{+0.004}_{-0.004}}$ & $0.645^{+0.006}_{-0.007}$ &  \\
 & 9 & $0.377^{+0.004}_{-0.007}$ & $0.463^{+0.008}_{-0.004}$ & $\mathbf{0.489^{+0.007}_{-0.004}}$ & $0.368^{+0.005}_{-0.009}$ &  \\
\midrule
$_\mathrm{upper}^\mathrm{v}$ & 1 & $0.923^{+0.002}_{-0.002}$ & $0.946^{+0.001}_{-0.001}$ & $\mathbf{0.950^{+0.001}_{-0.002}}$ & $0.928^{+0.002}_{-0.001}$ & \multirow{3}{*}{$\frac{\mathrm{m}}{\mathrm{s}}$} \\
 & 5 & $0.599^{+0.008}_{-0.010}$ & $\mathbf{0.736^{+0.006}_{-0.004}}$ & $\mathbf{0.742^{+0.006}_{-0.008}}$ & $0.596^{+0.008}_{-0.009}$ &  \\
 & 9 & $0.289^{+0.007}_{-0.010}$ & $0.397^{+0.003}_{-0.002}$ & $\mathbf{0.411^{+0.007}_{-0.011}}$ & $0.282^{+0.010}_{-0.008}$ &  \\
\midrule
$_\mathrm{upper}^\mathrm{z}$ & 1 & $0.985^{+0.001}_{-0.001}$ & $\mathbf{0.993^{+0.000}_{-0.000}}$ & $\mathbf{0.993^{+0.000}_{-0.001}}$ & $0.985^{+0.001}_{-0.001}$ & \multirow{3}{*}{$\mathrm{gpm}$} \\
 & 5 & $0.749^{+0.004}_{-0.006}$ & $\mathbf{0.867^{+0.004}_{-0.005}}$ & $\mathbf{0.870^{+0.004}_{-0.004}}$ & $0.741^{+0.006}_{-0.004}$ &  \\
 & 9 & $0.454^{+0.004}_{-0.008}$ & $0.572^{+0.004}_{-0.007}$ & $\mathbf{0.591^{+0.008}_{-0.005}}$ & $0.438^{+0.015}_{-0.021}$ &  \\
\bottomrule
\end{tabular}

\end{table}

\begin{figure}[h]
    \centering
    \includegraphics[width=0.45\linewidth, valign=t]{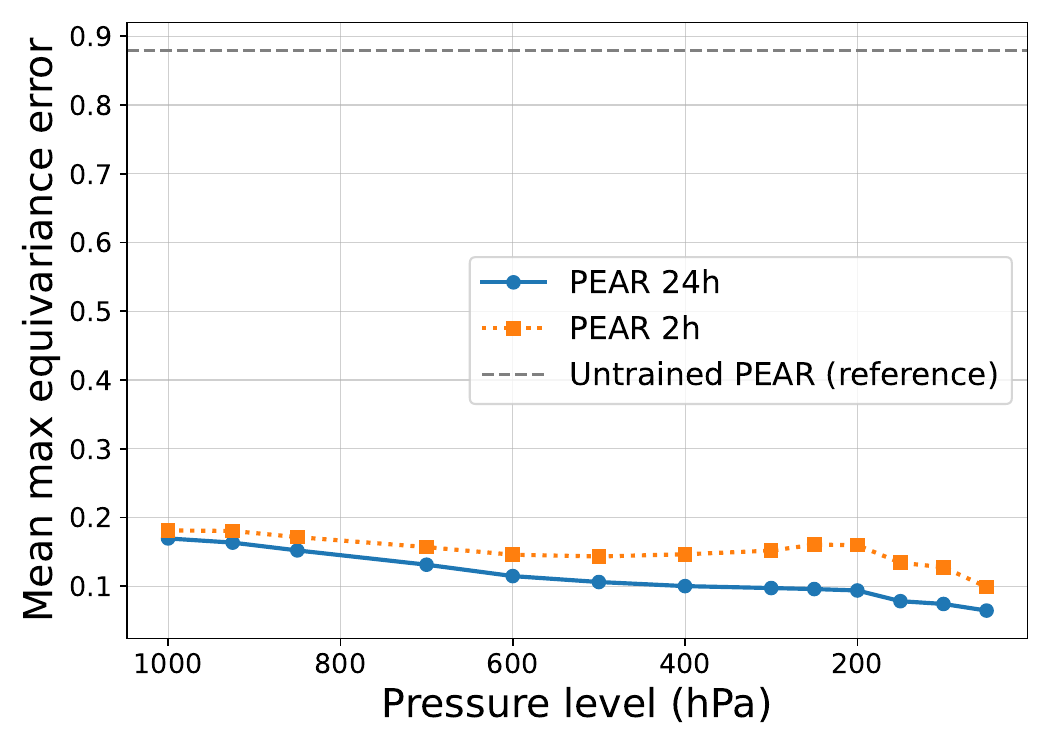}
    \hspace{0.9em}
    \includegraphics[width=0.45\linewidth, valign=t]{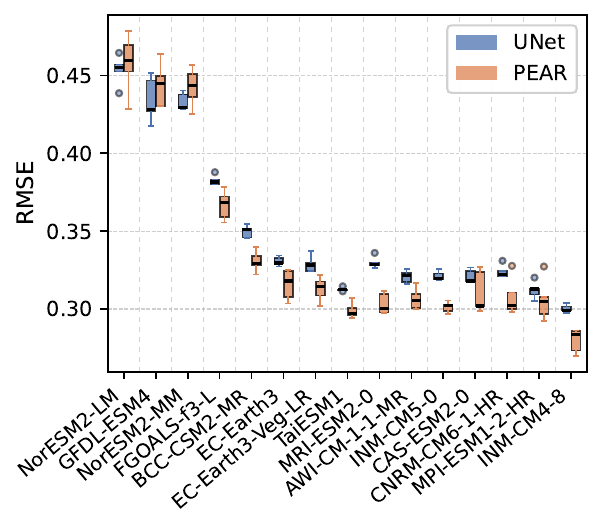}
    \caption{Left: Average of maximum equivariance error over rotation angles (see Appendix \ref{sec:appendix_equivariance} for details) over different pressure levels for the upper variables $q$, $t$, $u$, $v$ and $z$. PEAR 24h is trained on ERA5-Lite and PEAR 2h on a 2-hour sampling of 2012. Right: Overall RMSE for climate modelling over a number of different climate models \cite{kaltenborn2023climateset}.}
    \label{fig:equivariance_and_climate}
\end{figure}

\subsection{Equivariance}
Given the sparse time sampling of ERA5-lite with 24h time delta, we investigate if restricting to 00:00 UTC samples alters the equivariance properties of the weather prediction task. To this end, we use a new sampling of ERA5 at 2-hour intervals over one year. We calculate the equivariance error under rotations of the system around the polar axis as a function of the azimutal angle $\theta$. Figure~\ref{fig:equivariance_and_climate} shows the maximum equivariance error over $\theta$, averaged across upper-level variables, as a function of pressure level.
The blue and orange represent, respectively, PEAR trained on the 24-hour sampling and the 2-hour sampling datasets.
Both trained models lie well below the equivariance error of an untrained PEAR baseline (gray dashed line), showing a similar degree of learned equivariance regardless of the sampling strategy. PEAR trained on ERA5-Lite, spanning 2007-2017, achieves a slightly lower equivariance error than training on the 2-hour dataset restricted to 2012, indicating that ERA5-Lite is a good proxy for the full ERA5 dataset in terms of equivariance properties. Finally, on average, the equivariance error decreases with the altitude, in line with the expectation that landmasses break the rotational symmetry. However, not all the variables decrease monotonically; see appendix \ref{sec:appendix_equivariance} for details.

\subsection{Climate modeling}
To verify our architecture we also consider climate modeling, where the task is to approximate the outputs of computationally expensive Earth System Models (ESMs) across forcing scenarios at a fraction of the computational costs \cite{watson2021machine}. We train and evaluate PEAR on 15 models published in the ClimateSet dataset \cite{kaltenborn2023climateset} for five emission scenarios, using monthly emission forcings (CO2, SO2, CH4, BC) as inputs and surface temperature and precipitation as outputs. We compare PEAR against a UNet baseline~\cite{kaltenborn2023climateset}, where both models are evaluated on held-out emission scenarios across 15 climate models and 5 seeds. For PEAR the data is discretized onto HEALPix with  $n_{\mathrm{nside}}=32$. Figure~\ref{fig:equivariance_and_climate} shows the mean RMSE per climate model, averaged over output variables, for UNet and PEAR. PEAR achieves lower overall RMSE than UNet in 12 of the 15 climate models, outperforming UNet in precipitation predictions and showing comparable performance for temperature. See appendix \ref{sec:appendix_climate} for further details.

\section{Limitations}
\label{sec:limitations}
Our limited compute budget restricts us to ERA5-lite, making comparisons to models trained on the full ERA5 dataset harder. Ideally we would train on ground truth data created on the HEALPix grid, but at this time the reanalysis for ERA5 is done on an equiangular grid.
Known limitations of data-driven forecasts include unphysical predictions~\cite{schreck2024}, instabilities for longer forecasts~\cite{watt-meyer2024} and lack of certain features such as the butterfly effect~\cite{selz2023} or sub-synoptic and mesoscale weather phenomena~\cite{bonavita2024}. Although we have not checked this explicitly, we suspect that our model is also subject to these limitations.

\section{Conclusion}

We have shown the importance of using the HEALPix grid that minimizes unphysical biases for global medium term weather prediction. The hierarchical equal area pixelation enables efficient implementation of our transformer architecture PEAR, that outperforms its counterpart on the traditional equiangular grid at no computational overhead. With forecast horizons of up to 10 days, we showed that PEAR outperforms the equiangular baseline with more than twice the number of trainable parameters. The superior performance of PEAR should also lead to more accurate extreme weather forecasting~\cite{olivetti2024}, a direction that would be interesting for future work.

As next generation sources for data driven weather and climate forecasting aim to include high resolution HEALPix native data~\cite{grib1togrib2}, we hope that PEAR can pave the way for using this data in the most efficient manner.

\section*{Acknowledgments}
H.L. thanks Karl Bergström for advice and support in developing custom software for model tracking and interactive visualization of spherical data.
The work of J.G. and D.P. is supported by the Wallenberg AI, Autonomous Systems and Software Program (WASP) funded by the Knut and Alice Wallenberg (KAW) Foundation. The computations were enabled by resources provided by the National Academic Infrastructure for Supercomputing in Sweden (NAISS),
partially funded by the Swedish Research Council through grant agreement no. 2022-06725.

\renewcommand*{\bibfont}{\normalfont\footnotesize}
\printbibliography

\newpage
\appendix

\label{appendix}
\setcounter{figure}{0} 
\setcounter{table}{0}
\numberwithin{figure}{section}
\numberwithin{table}{section}
\numberwithin{equation}{section}

\section{PEAR}

\begin{table}
\centering
\caption{Average root mean squared error (RMSE) for three different prediction lead times (1, 3 and 5 days). Bold face indicates best model. Mean $\pm$ std across ensemble members where available.}
\label{tab:rmse}
\small
\renewcommand{\arraystretch}{1.4}
\begin{tabular}{l r r r r r l}
\toprule
Variable & $\Delta t$ & \multicolumn{4}{c}{RMSE} & unit \\
\cmidrule(lr){3-6}
 & (days) & FengWu & GraphCast & PEAR & PanguPN & \\
\midrule
$_\mathrm{surface}^\mathrm{msl}$ & 1 & $99^{+2}_{-3}$ & $\mathbf{76^{+1}_{-1}}$ & $\mathbf{76^{+4}_{-2}}$ & $100^{+2}_{-3}$ & \multirow{3}{*}{$\mathrm{Pa}$} \\
 & 5 & $347^{+8}_{-9}$ & $\mathbf{271^{+11}_{-6}}$ & $\mathbf{265^{+6}_{-5}}$ & $349^{+6}_{-7}$ &  \\
 & 9 & $495^{+13}_{-15}$ & $\mathbf{462^{+6}_{-6}}$ & $\mathbf{450^{+10}_{-8}}$ & $500^{+7}_{-7}$ &  \\
\midrule
$_\mathrm{surface}^\mathrm{t2m}$ & 1 & $0.85^{+0.02}_{-0.01}$ & $0.74^{+0.01}_{-0.01}$ & $\mathbf{0.71^{+0.01}_{-0.01}}$ & $0.88^{+0.01}_{-0.02}$ & \multirow{3}{*}{$\mathrm{K}$} \\
 & 5 & $1.54^{+0.09}_{-0.06}$ & $1.40^{+0.08}_{-0.05}$ & $\mathbf{1.27^{+0.07}_{-0.04}}$ & $1.57^{+0.02}_{-0.03}$ &  \\
 & 9 & $2.05^{+0.20}_{-0.13}$ & $2.05^{+0.13}_{-0.12}$ & $\mathbf{1.79^{+0.11}_{-0.06}}$ & $2.07^{+0.04}_{-0.04}$ &  \\
\midrule
$_\mathrm{surface}^\mathrm{u10}$ & 1 & $1.12^{+0.02}_{-0.01}$ & $\mathbf{0.96^{+0.01}_{-0.01}}$ & $\mathbf{0.95^{+0.02}_{-0.01}}$ & $1.14^{+0.02}_{-0.02}$ & \multirow{3}{*}{$\frac{\mathrm{m}}{\mathrm{s}}$} \\
 & 5 & $2.43^{+0.02}_{-0.03}$ & $2.12^{+0.06}_{-0.03}$ & $\mathbf{2.06^{+0.03}_{-0.02}}$ & $2.43^{+0.05}_{-0.04}$ &  \\
 & 9 & $3.10^{+0.06}_{-0.04}$ & $3.03^{+0.05}_{-0.03}$ & $\mathbf{2.96^{+0.03}_{-0.03}}$ & $3.15^{+0.05}_{-0.06}$ &  \\
\midrule
$_\mathrm{surface}^\mathrm{v10}$ & 1 & $1.15^{+0.03}_{-0.01}$ & $\mathbf{0.99^{+0.01}_{-0.01}}$ & $\mathbf{0.97^{+0.02}_{-0.01}}$ & $1.18^{+0.02}_{-0.02}$ & \multirow{3}{*}{$\frac{\mathrm{m}}{\mathrm{s}}$} \\
 & 5 & $2.54^{+0.02}_{-0.02}$ & $\mathbf{2.20^{+0.05}_{-0.03}}$ & $\mathbf{2.14^{+0.04}_{-0.02}}$ & $2.55^{+0.04}_{-0.04}$ &  \\
 & 9 & $3.23^{+0.06}_{-0.03}$ & $\mathbf{3.18^{+0.07}_{-0.04}}$ & $\mathbf{3.12^{+0.03}_{-0.03}}$ & $3.28^{+0.05}_{-0.03}$ &  \\
\midrule
$_\mathrm{upper}^\mathrm{q}$ & 1 & $3.00^{+0.02}_{-0.02}$ & $\mathbf{2.68^{+0.03}_{-0.03}}$ & $\mathbf{2.62^{+0.04}_{-0.02}}$ & $3.14^{+0.03}_{-0.03}$ & \multirow{3}{*}{$\frac{\mathrm{g}}{\mathrm{kg}}$ $\times 10^{-4}$} \\
 & 5 & $4.89^{+0.06}_{-0.04}$ & $4.36^{+0.05}_{-0.04}$ & $\mathbf{4.23^{+0.07}_{-0.06}}$ & $4.92^{+0.16}_{-0.07}$ &  \\
 & 9 & $6.02^{+0.08}_{-0.09}$ & $\mathbf{5.75^{+0.04}_{-0.05}}$ & $\mathbf{5.62^{+0.08}_{-0.13}}$ & $6.03^{+0.28}_{-0.10}$ &  \\
\midrule
$_\mathrm{upper}^\mathrm{t}$ & 1 & $0.82^{+0.02}_{-0.01}$ & $0.72^{+0.01}_{-0.01}$ & $\mathbf{0.68^{+0.01}_{-0.01}}$ & $0.83^{+0.02}_{-0.01}$ & \multirow{3}{*}{$\mathrm{K}$} \\
 & 5 & $1.78^{+0.06}_{-0.03}$ & $1.51^{+0.04}_{-0.03}$ & $\mathbf{1.40^{+0.02}_{-0.01}}$ & $1.80^{+0.06}_{-0.03}$ &  \\
 & 9 & $2.45^{+0.11}_{-0.06}$ & $2.27^{+0.06}_{-0.04}$ & $\mathbf{2.14^{+0.02}_{-0.02}}$ & $2.46^{+0.09}_{-0.06}$ &  \\
\midrule
$_\mathrm{upper}^\mathrm{u}$ & 1 & $2.30^{+0.03}_{-0.02}$ & $1.98^{+0.02}_{-0.03}$ & $\mathbf{1.86^{+0.03}_{-0.02}}$ & $2.22^{+0.03}_{-0.03}$ & \multirow{3}{*}{$\frac{\mathrm{m}}{\mathrm{s}}$} \\
 & 5 & $4.95^{+0.02}_{-0.03}$ & $4.24^{+0.13}_{-0.08}$ & $\mathbf{4.04^{+0.04}_{-0.03}}$ & $4.95^{+0.09}_{-0.07}$ &  \\
 & 9 & $6.67^{+0.06}_{-0.04}$ & $6.31^{+0.09}_{-0.09}$ & $\mathbf{6.18^{+0.03}_{-0.09}}$ & $6.66^{+0.09}_{-0.07}$ &  \\
\midrule
$_\mathrm{upper}^\mathrm{v}$ & 1 & $2.28^{+0.03}_{-0.03}$ & $1.94^{+0.02}_{-0.03}$ & $\mathbf{1.85^{+0.03}_{-0.02}}$ & $2.21^{+0.02}_{-0.03}$ & \multirow{3}{*}{$\frac{\mathrm{m}}{\mathrm{s}}$} \\
 & 5 & $4.93^{+0.07}_{-0.04}$ & $\mathbf{4.14^{+0.09}_{-0.07}}$ & $\mathbf{4.05^{+0.08}_{-0.03}}$ & $4.92^{+0.07}_{-0.07}$ &  \\
 & 9 & $6.47^{+0.09}_{-0.07}$ & $\mathbf{6.22^{+0.08}_{-0.06}}$ & $\mathbf{6.17^{+0.10}_{-0.08}}$ & $6.45^{+0.09}_{-0.08}$ &  \\
\midrule
$_\mathrm{upper}^\mathrm{z}$ & 1 & $96^{+4}_{-3}$ & $\mathbf{74^{+2}_{-2}}$ & $\mathbf{70^{+2}_{-1}}$ & $97^{+3}_{-4}$ & \multirow{3}{*}{$\mathrm{gpm}$} \\
 & 5 & $361^{+10}_{-5}$ & $282^{+9}_{-8}$ & $\mathbf{265^{+3}_{-3}}$ & $373^{+4}_{-8}$ &  \\
 & 9 & $547^{+12}_{-9}$ & $506^{+11}_{-5}$ & $\mathbf{483^{+4}_{-8}}$ & $563^{+13}_{-7}$ &  \\
\bottomrule
\end{tabular}

\end{table}

\begin{table}
\centering
\renewcommand{\arraystretch}{1.4}
\caption{PEAR architecture overview. The windowed attention blocks contain multiple copies, indicated by the depth column, of the multi-head attention layer and layer normalization. The tensor shape column shows the shape of the output tensor from the corresponding layer. Both the input and final patch recovery layer use two separate tensors for the surface and upper variables.
    \label{tab:arch}}
    \begin{tabular}{clcll}
        \toprule
         Nr & Layer/Block & Depth & Tensor shape & Attention heads \\
         \midrule 
         &Input& 1 &  $\left\{\begin{array}{l}(12 n_{\mathrm{side}}^2, 4) \\
          (12 n_{\mathrm{side}}^2, 13, 5)\end{array}\right.$ \\
         1&Patch embed & 1 & $(\nicefrac{3}{4} n_{\mathrm{side}}^2, 8, 48)$   \\
         2&Windowed attention & 2 & $(\nicefrac{3}{4} n_{\mathrm{side}}^2, 8, 48)$ & 6   \\
         3&Downsample & 1 & $(\nicefrac{3}{16}n_{\mathrm{side}}^2, 8, 96)$   \\
         4&Windowed attention & 12 & $(\nicefrac{3}{16}n_{\mathrm{side}}^2, 8, 96)$ & 12  \\
         5&Upsample & 1 & $(\nicefrac{3}{4} n_{\mathrm{side}}^2, 8, 48)$   \\
         6&Windowed attention & 2 & $(\nicefrac{3}{4} n_{\mathrm{side}}^2, 8, 48)$ & 6   \\
         & Concatenate 6 \& 2 & 1 & $(\nicefrac{3}{4} n_{\mathrm{side}}^2, 8, 96)$ \\
         & Patch recovery & 1 & 
        $\left\{\begin{array}{l}(12 n_{\mathrm{side}}^2, 4) \\
          (12 n_{\mathrm{side}}^2, 13, 5)\end{array}\right.$ \\
          \bottomrule
    \end{tabular}
\end{table}

\begin{figure}[t]
    \centering
    \includegraphics[width=0.35\linewidth]{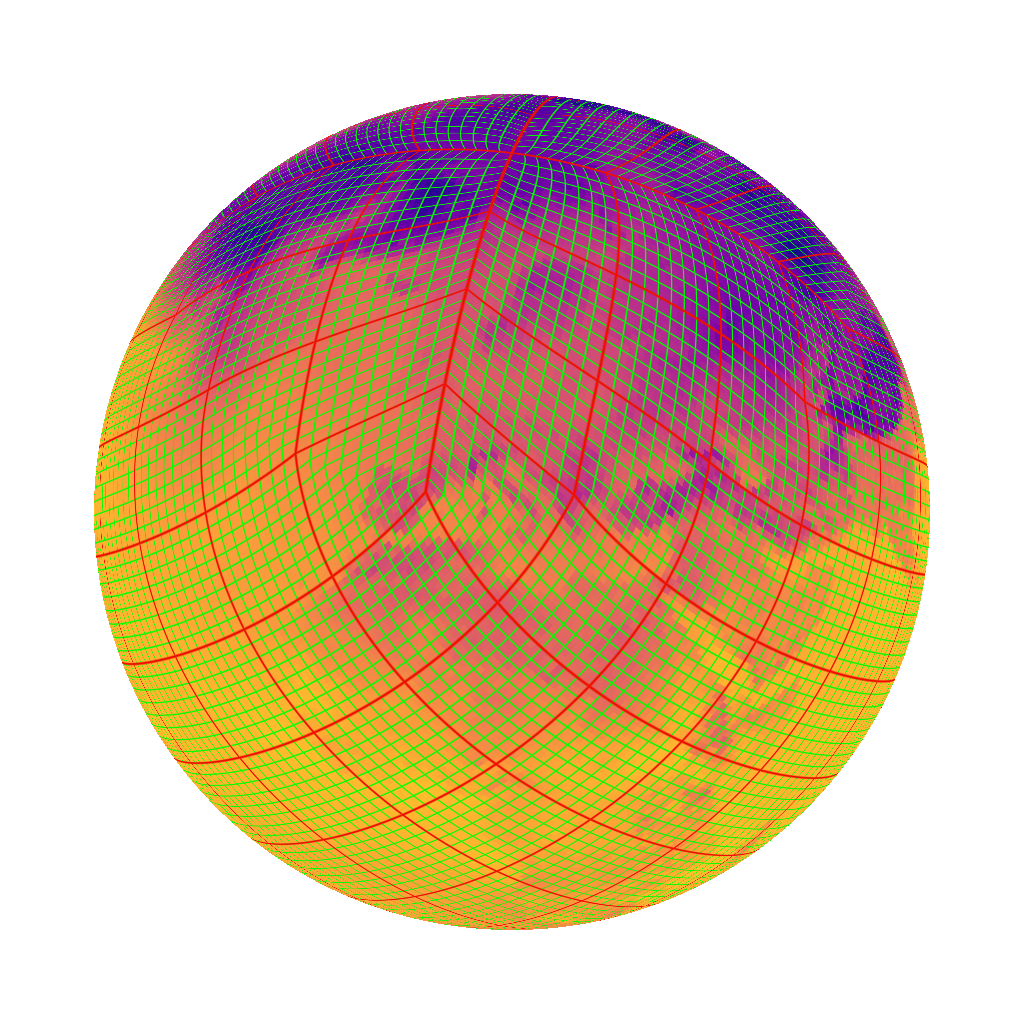}
    \caption{Patches (green) and windows (red) at the first attention layer. Each green patch contains four pixels of the input resolution $n_\mathrm{side}=64$ on the HEALPix grid, and each window contains 64 patches at this level. The window also include one vertical level above, not shown in the figure.}
    \label{fig:patch_and_window}
\end{figure}

An overview of the architecture of PEAR can be found in Table. \ref{tab:arch}. The patches and windows are visualized in Figure. \ref{fig:patch_and_window}. Initial patch embed over 16 nested HEALPix pixels, and all windows are $[2, 64]$ in [level, healpix].

We use RMSE and Anomaly Correlation Coefficient (ACC) calculated on the HEALPix grid according to
\begin{equation}
    \mathrm{RMSE}(y, \hat{y}) = \sqrt{\frac{1}{12 n_{\mathrm{side}}^2}\sum_{i=0}^{12 n_{\mathrm{side}}^2} \left( y^i - \hat{y}^i\right)^2}
    \label{eq:rmse}
\end{equation}

\begin{equation}
    \mathrm{ACC}(y, \hat{y}) = \frac{\sum_{i}^{12n^2_\mathrm{nside}}\Delta y^i \Delta \hat{y}^i}{\sqrt{\left(\sum_{i}^{12n^2_\mathrm{nside}}(\Delta y^i)^2\right)\left(\sum_{i}^{12n^2_\mathrm{nside}}(\Delta \hat{y}^i)^2\right)}}\,,
    \label{eq:acc}
\end{equation}
where $\Delta y$ is the difference between the predictions and the climatology average.

\begin{figure}
    \centering
    \includegraphics[width=0.9\linewidth,height=0.85\textheight,keepaspectratio]{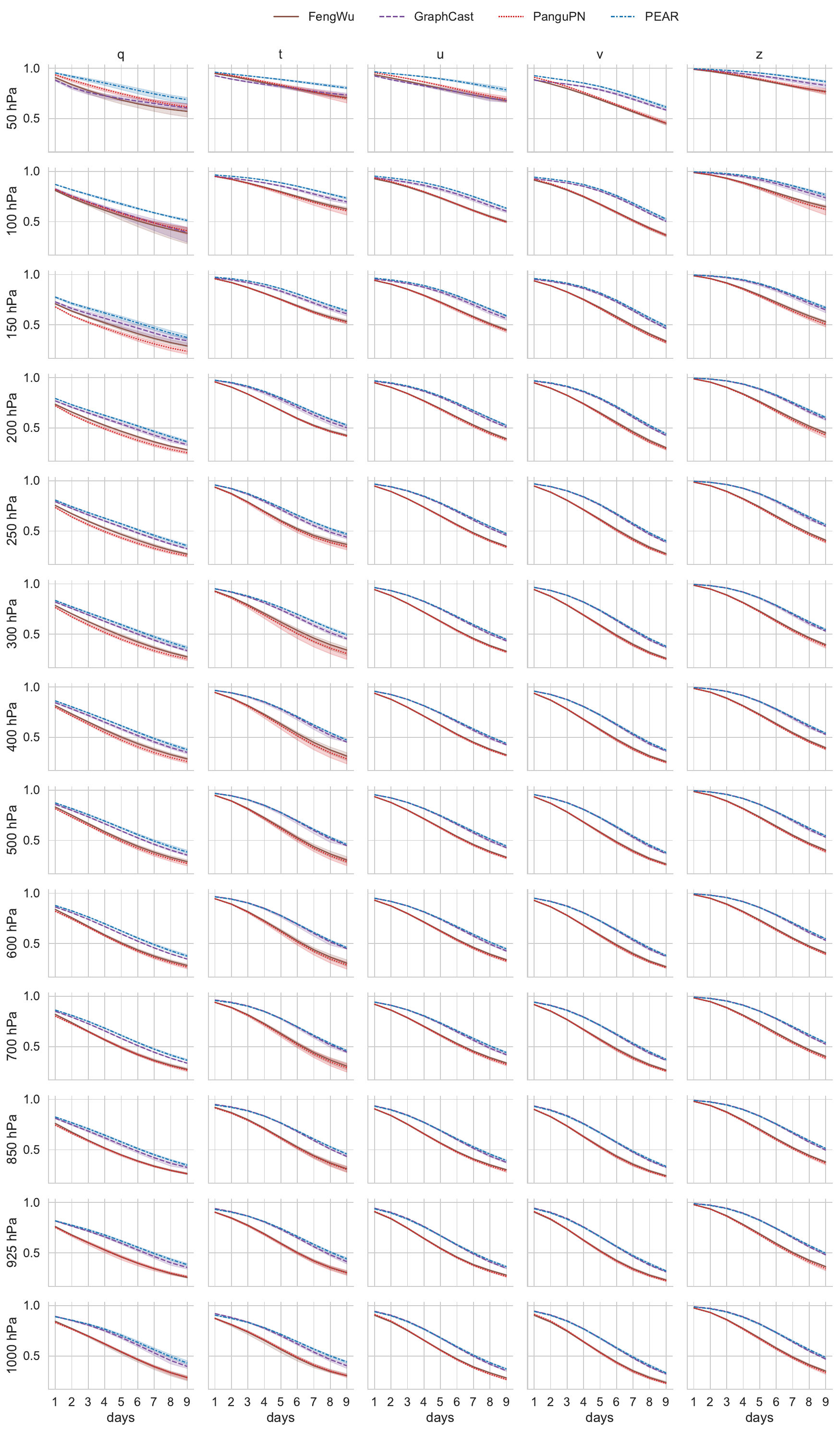}
    \caption{Anomaly correlation coefficient (ACC) for the upper variables separated by pressure level over up to 10 days lead time.}
    \label{fig:acc_iterated_upper}
\end{figure}

\begin{figure}
    \centering
    \includegraphics[width=0.9\linewidth,height=0.85\textheight,keepaspectratio]{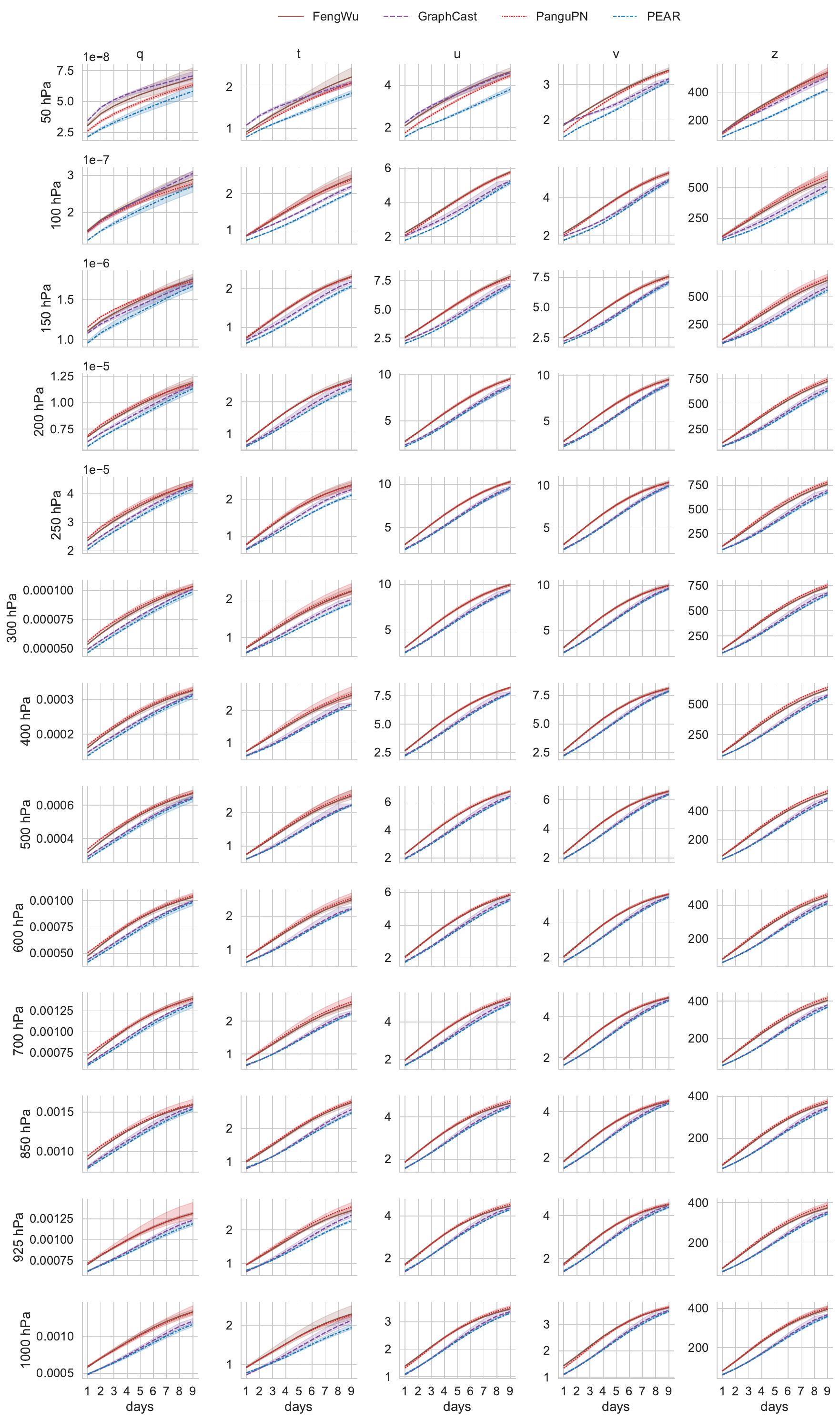}
    \caption{Root mean squared error (RMSE) for the upper variables separated by pressure level over up to 10 days lead time.}
    \label{fig:rmse_iterated_upper}
\end{figure}

We include anomaly correlation coefficient (ACC) and root mean squared error (RMSE) for all upper variables and pressure levels in Fig.~\ref{fig:acc_iterated_upper} and Fig.~\ref{fig:rmse_iterated_upper} respectively. PEAR outperforms Pangu for all variables and pressure levels for ACC in Fig.~\ref{fig:acc_iterated_upper}, and often outperforms the almost eight times larger PanguLarge.

\begin{figure}[h]
    \centering
    \includegraphics[width=\textwidth,height=0.85\textheight,keepaspectratio]{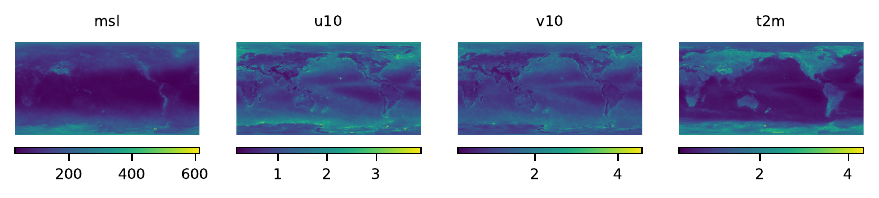}
    \caption{Spatial RMSE for PEAR predictions of surface variables with one day lead time averaged over the validation year. The HEALPix predictions are projected to a cartesian longitude (horizontal) and latitude (vertical) grid for visualization.}
    \label{fig:spatial_surface_pear}
\end{figure}

\begin{figure}[h]
    \centering
    \includegraphics[width=\textwidth,height=0.85\textheight,keepaspectratio]{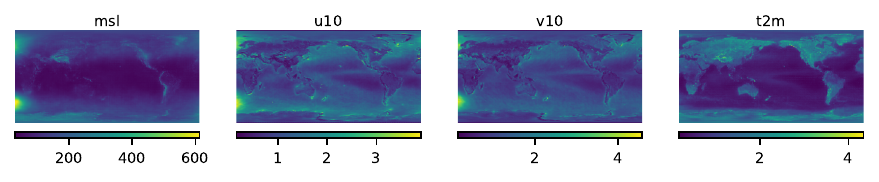}
    \caption{Spatial RMSE for Pangu predictions of surface variables with one day lead time averaged over the validation year.}
    \label{fig:spatial_surface_pangu}
\end{figure}

\begin{figure}
    \centering
    \includegraphics[width=\textwidth,height=0.85\textheight,keepaspectratio]{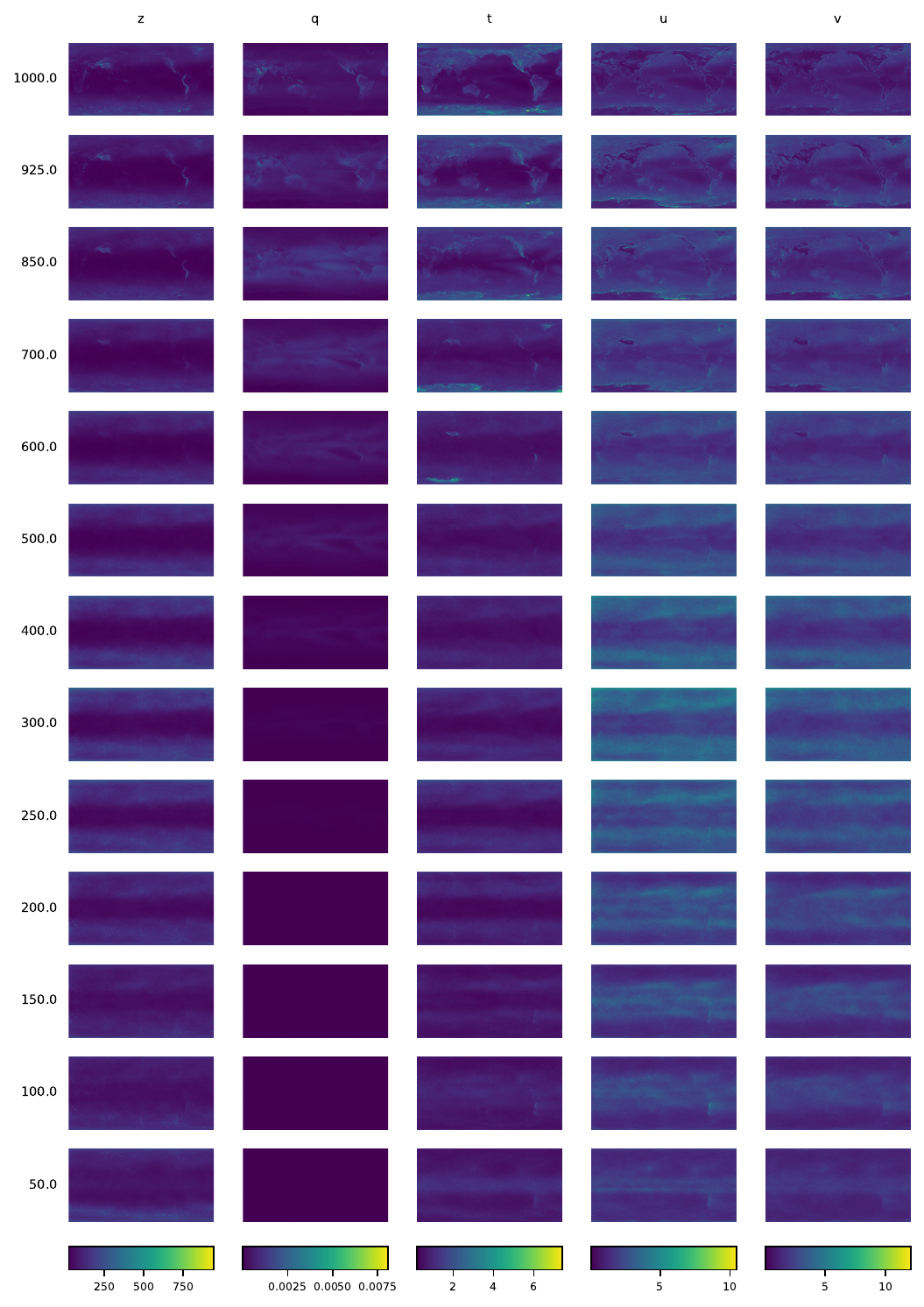}
    \caption{Spatial RMSE for PEAR predictions of upper variables with one day lead time averaged over the validation year. Pressure level (rows) and variables (columns) with joint color mapping per variable. The HEALPix predictions are projected to a cartesian longitude (horizontal) and latitude (vertical) grid for visualization.}
    \label{fig:spatial_pear_upper}
\end{figure}

\begin{figure}
    \centering
    \includegraphics[width=\textwidth,height=0.85\textheight,keepaspectratio]{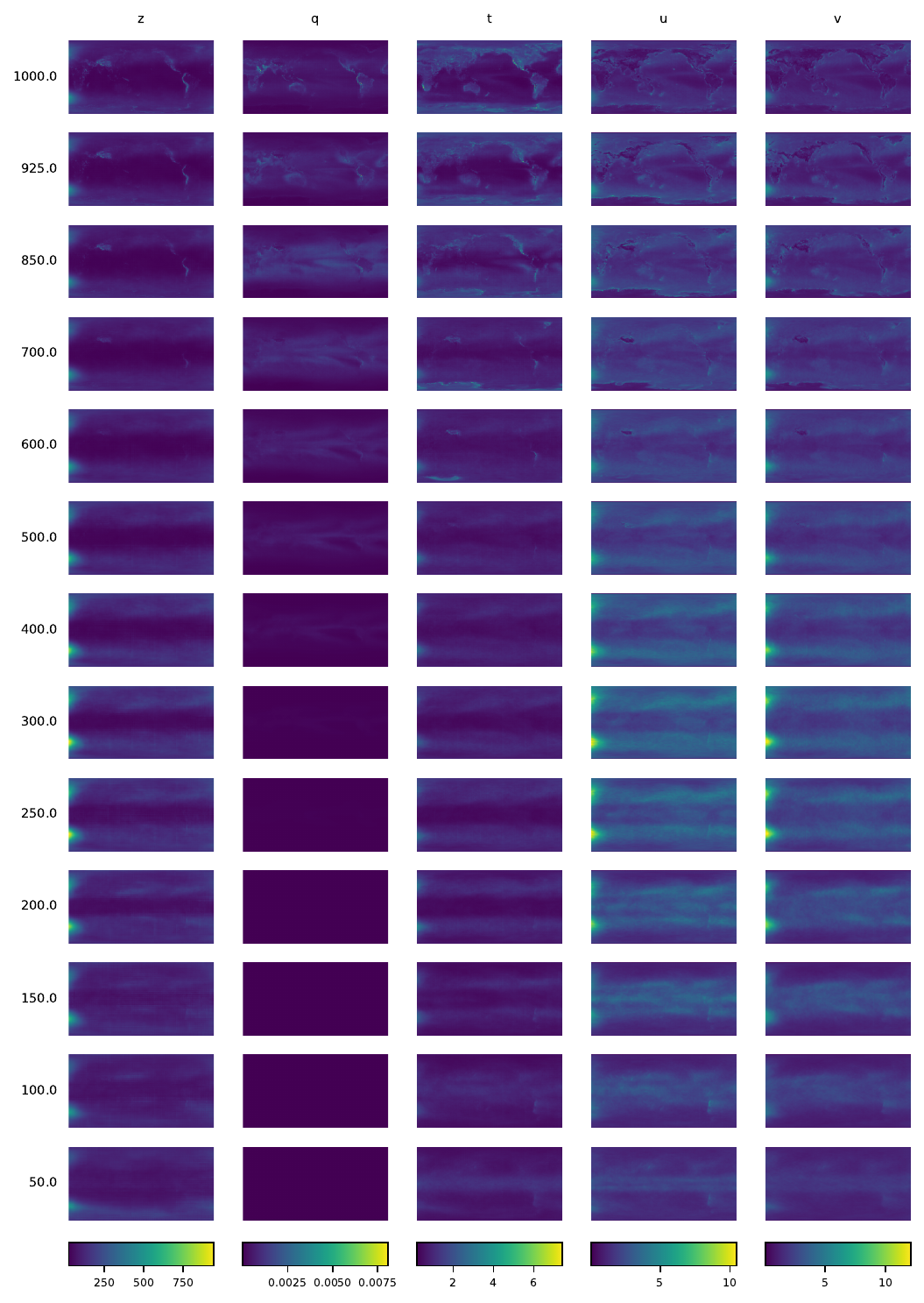}
    \caption{Spatial RMSE for Pangu predictions of upper variables with one day lead time averaged over the validation year. Pressure level (rows) and variables (columns) with joint color mapping per variable.}
    \label{fig:spatial_pangu_upper}
\end{figure}

We also include spatially resolved RMSE over the validation year for \includegraphics[height=1em]{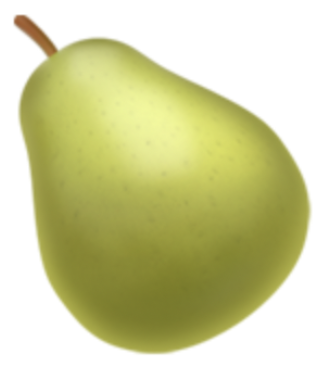} in Fig.~\ref{fig:spatial_surface_pear} (surface) and~\ref{fig:spatial_pear_upper} (upper), as well as the corresponding spatial RMSE for Pangu in Fig.~\ref{fig:spatial_surface_pangu} (surface) and Fig.\ref{fig:spatial_pangu_upper} (upper). 

\begin{figure}[p]
    \centering
    \includegraphics[width=\textwidth,height=0.9\textheight,keepaspectratio]{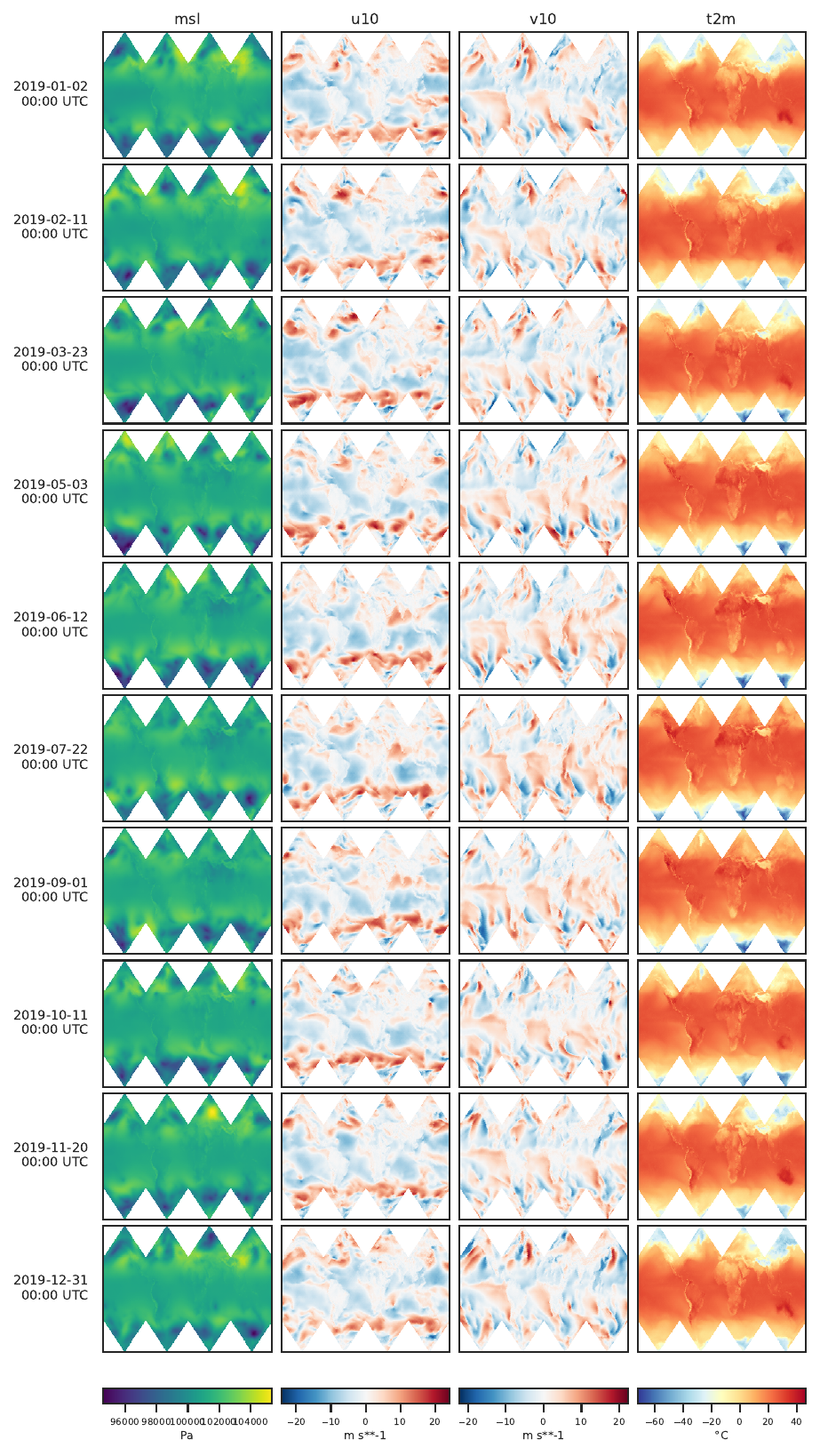}
    \caption{PEAR surface variable predictions with one day lead time.}
    \label{fig:pear_epoch0200_surface}
\end{figure}

\begin{figure}[p]
    \centering
    \includegraphics[width=\textwidth,height=0.9\textheight,keepaspectratio]{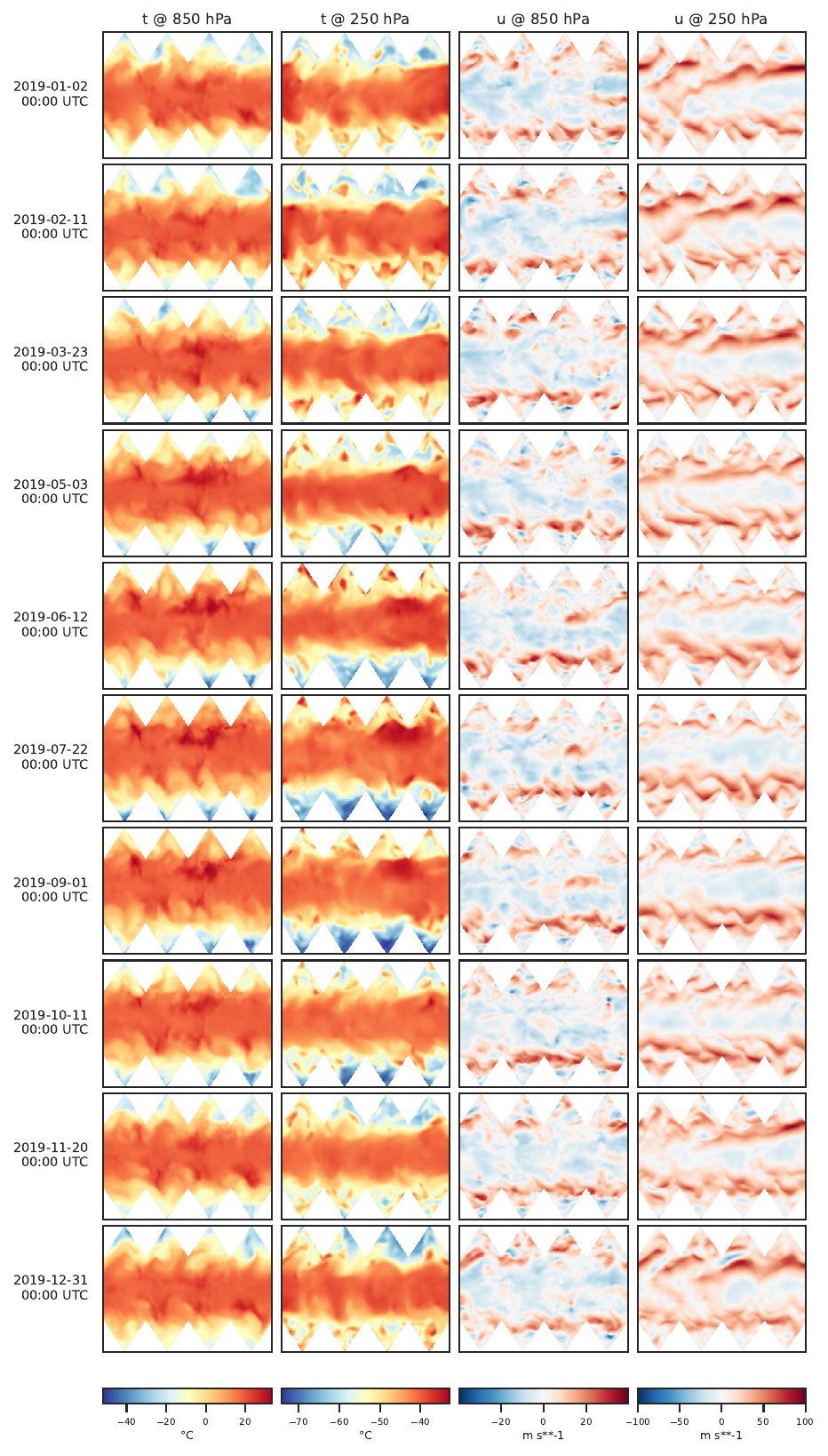}
    \caption{PEAR upper variable predictions with one day lead time. Two pressure levels (250 and 850) and two variables (temperature and wind north-south) are shown.}
    \label{fig:pear_epoch0200_upper}
\end{figure}
\section{Climate}
\label{sec:appendix_climate}

A related but distinct challenge that has recently attracted machine learning interest is climate model emulation. Unlike medium-range weather forecasting, where the task is to predict the future atmospheric state from initial conditions, emulation aims to approximate the outputs of computationally intensive Earth System Models (ESMs) for a given set of inputs. ESMs require substantial computational resources and individual runs can take months, severely limiting the number of scenarios that can be explored \cite{watson2021machine}. To support machine learning research in this area, several dedicated datasets have been developed such as ClimateBench \cite{watson2022climatebench}, ClimateLearn \cite{nguyen2023climatelearn}, Climart \cite{cachay2021climart} and ClimateSet \cite{kaltenborn2023climateset}. On the modeling side, a range of architectures have been explored, including Convolutional LSTMs \cite{watson2022climatebench}, UNet models \cite{kaltenborn2023climateset} and transformer architectures such as ClimaX \cite{nguyen2023a}. %

PEAR has been evaluated on climate model emulation using the ClimateSet dataset, a large-scale, machine-learning-ready dataset containing processed outputs for 21 different climate models from the CMIP6 ensemble paired with a shared set of forcing inputs (Input4MIPS). The input dataset is a processed collection of monthly-frequency emission maps for four primary climate forcing agents (CO2, SO2, CH4, and BC), and the output variables are global, monthly averaged maps of surface air temperature and precipitation at a resolution of 2.5\textdegree × 1.875\textdegree — significantly coarser than the ERA5 data on which the original PEAR model was trained. The data covers a historical scenario (1850–2014) and four future Shared Socioeconomic Pathways (SSPs): SSP1-2.6, SSP2-4.5, SSP3-7.0, and SSP5-8.5, covering the period 2014–2100. For use with PEAR, the dataset has been discretized onto the HEALPix grid with $n_{\mathrm{nside}}=32$.

ClimateSet also provides benchmark models for comparison: a UNet, a ConvLSTM, and a ClimaX implementation. Of these, only the ConvLSTM explicitly models temporal correlations across time steps, while the UNet and ClimaX, as implemented by ClimateSet, treat each time step independently. The initial PEAR implementation for climate also uses a single input time step, deferring temporal context to later experiments. The ConvLSTM is therefore excluded from comparison to keep baselines aligned.

The PEAR architecture has been adapted to handle the single set of surface-level fields, with 4 input channels for the climate forcing agents. The architecture is simplified to a single-branch patch embedding using only a 1D convolution, collapsing the depth dimension to $D = 1$ throughout all transformer stages. A single output head then projects back to the desired number of output channels. In this configuration, the model still has 4.3M trainable parameters.

We evaluate PEAR against two baselines from the ClimateSet benchmark, a UNet with 18.3M trainable parameters and a ClimaX model with 106.6M trainable parameters. Each model is trained separately from scratch on 15 of the 21 available climate models with 5 random seeds, and evaluated on the held-out test set of the corresponding climate model. Training follows the original ClimateSet experimental setup, using the SSP1-2.6, SSP3-7.0, and SSP5-8.5 scenarios with 10\% of the data held out for validation, and testing on the SSP2-4.5 scenario. The reported metric is the RMSE computed on normalized labels, averaged over temperature and precipitation.

\begin{figure}[p]
    \centering
    \begin{minipage}{\textwidth}
        \centering
        \includegraphics[width=\textwidth]{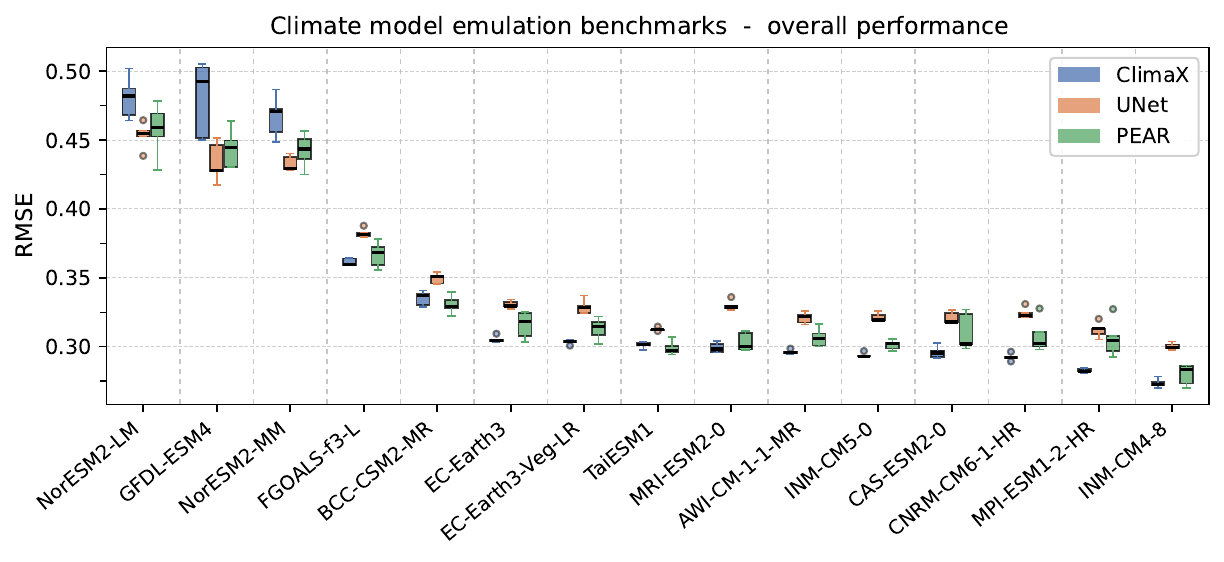}
    \end{minipage}

    \vspace{0.3cm}

    \begin{minipage}{\textwidth}
        \centering
        \includegraphics[width=\textwidth]{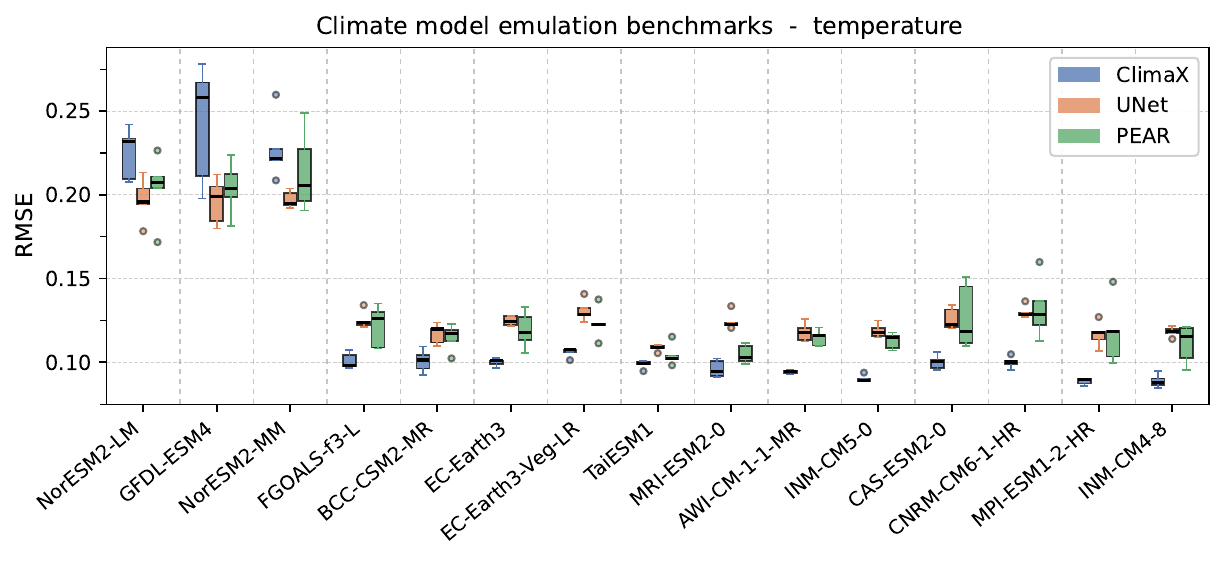}
    \end{minipage}

    \vspace{0.3cm}

    \begin{minipage}{\textwidth}
        \centering
        \includegraphics[width=\textwidth]{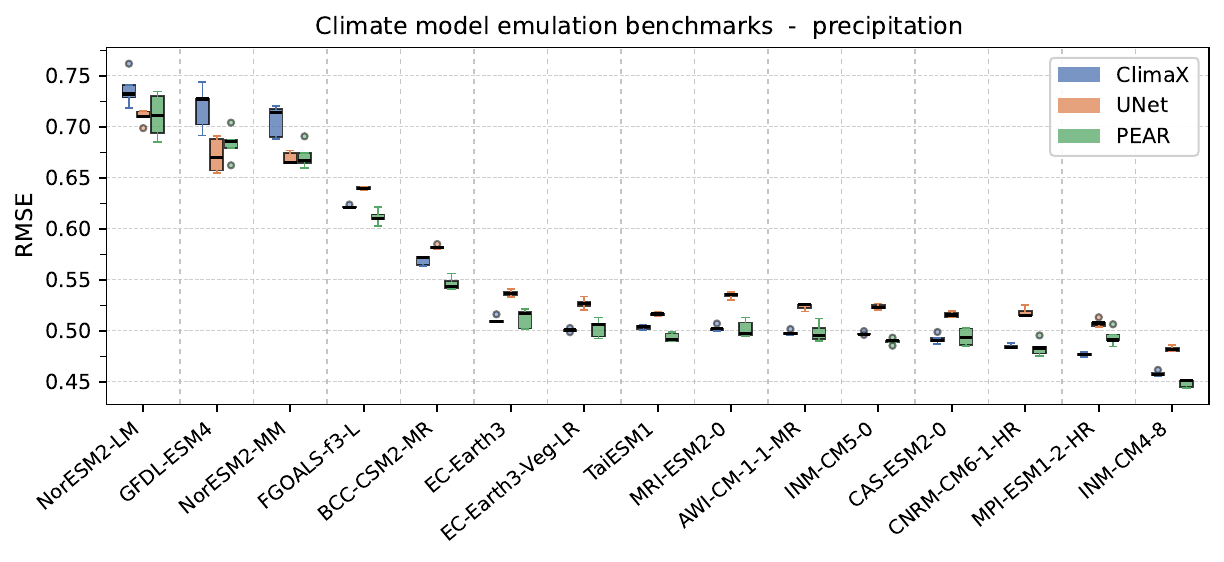}
    \end{minipage}
    \caption{Benchmarking results for climate model emulation. Box plots showing RMSE on normalized outputs for ClimaX, UNet, and PEAR across 15 ClimateSet climate models. Plot from top to bottom showing (a) Overall performance. (b) Surface air temperature. (c) Precipitation.}
    \label{fig:climate_appendix_plot}
\end{figure}

Figure \ref{fig:climate_appendix_plot} shows the per-variable and overall RMSE for each model across the 15 climate models and 5 seeds. In terms of overall RMSE, ClimaX achieves the lowest error on 10 of the 15 climate models, followed by UNet on 3 and PEAR on the remaining 2. Despite being roughly four times smaller than UNet, PEAR outperforms it on 13 of the 15 models in overall RMSE. Separating the results by variable, ClimaX performs best on surface temperature for 12 of the 15 climate models, with UNet performing best on the remaining 3. For precipitation, however, the trend differs with PEAR achieving the lowest RMSE on 8 of the 15 climate models, compared with 4 for ClimaX and 3 for UNet.

It is unsurprising that ClimaX outperforms the other models given its considerably larger parameter count. More notable is that PEAR generally matches or exceeds UNet across both variables despite being roughly four times smaller, and achieves the lowest precipitation RMSE for a majority of the climate models. Although PEAR does not achieve the same relative advantage for surface temperature as it does for precipitation, its strong performance relative to its size leaves open the possibility that further experimentation with model capacity and architectural refinements could improve results on temperature as well. Overall, these results demonstrate that PEAR is well-suited for the task of climate model emulation, achieving competitive performance despite having a substantially smaller parameter count. \\

\section{Equivariance}
\label{sec:appendix_equivariance}

The ERA5-Lite dataset uses samples at 00:00 UTC between 2007 and 2017 (4017 samples) from the full ERA5 dataset which breaks the rotational symmetry introduced by the rotating illumination of the sun. In order to investigate how this effect influences the symmetry properties of the weather prediction task and therefore of the models we train, we compare PEAR trained on ERA5-Lite to the same model trained on a new dataset, sampled from the full ERA5 dataset at 2-hour intervals from 2012, resulting in 4380 samples. On both datasets, models are trained to evolve the state of the atmosphere by 24h and we use the year 2019 for evaluation, with the same sampling strategy used as for the respective training data.

In the following, we will investigate the equivariance properties of the trained models in order to analyze the equivariance of the learning tasks defined by the two datasets. To this end, we measure the equivariance error under azimuthal rotations of the sphere.
Therefore, we consider the group $SO(2)$ of rotations, parametrized by an angle $\theta \in [0,2\pi)$. Let $R_\theta$ denote the rotation by the angle $\theta$ around the polar axis, and let $f: \mathcal{X} \rightarrow \mathcal{Y}$ represent the model that maps the input space $\mathcal{X}$ to the output space $\mathcal{Y}$. In this case, a mapping $f$ is said to be \textit{equivariant} if the rotation of the input commutes with the rotation of the output as follows:
\[
f(R_\theta\mathbf{x}) = R_\theta f(\mathbf{x}), \;\;\forall \mathbf{x} \in \mathcal{X}, \forall \theta \in [0,2\pi)
\]
Therefore, to quantify the degree to which $f$ satisfies this property, we define the \textit{mean equivariance error}:
\[
\mathcal{E}(f,\theta) = \mathbb{E}_{\mathbf{x}\sim\mathcal{X}}[|f(R_\theta\mathbf{x}) - R_\theta f(\mathbf{x})|]
\]
where the expectation is taken over the data distribution of the validation set. %
The error is computed individually for each variable, since each feature encodes distinct information.
\begin{figure}
    \centering
    \includegraphics[width=0.6\linewidth]{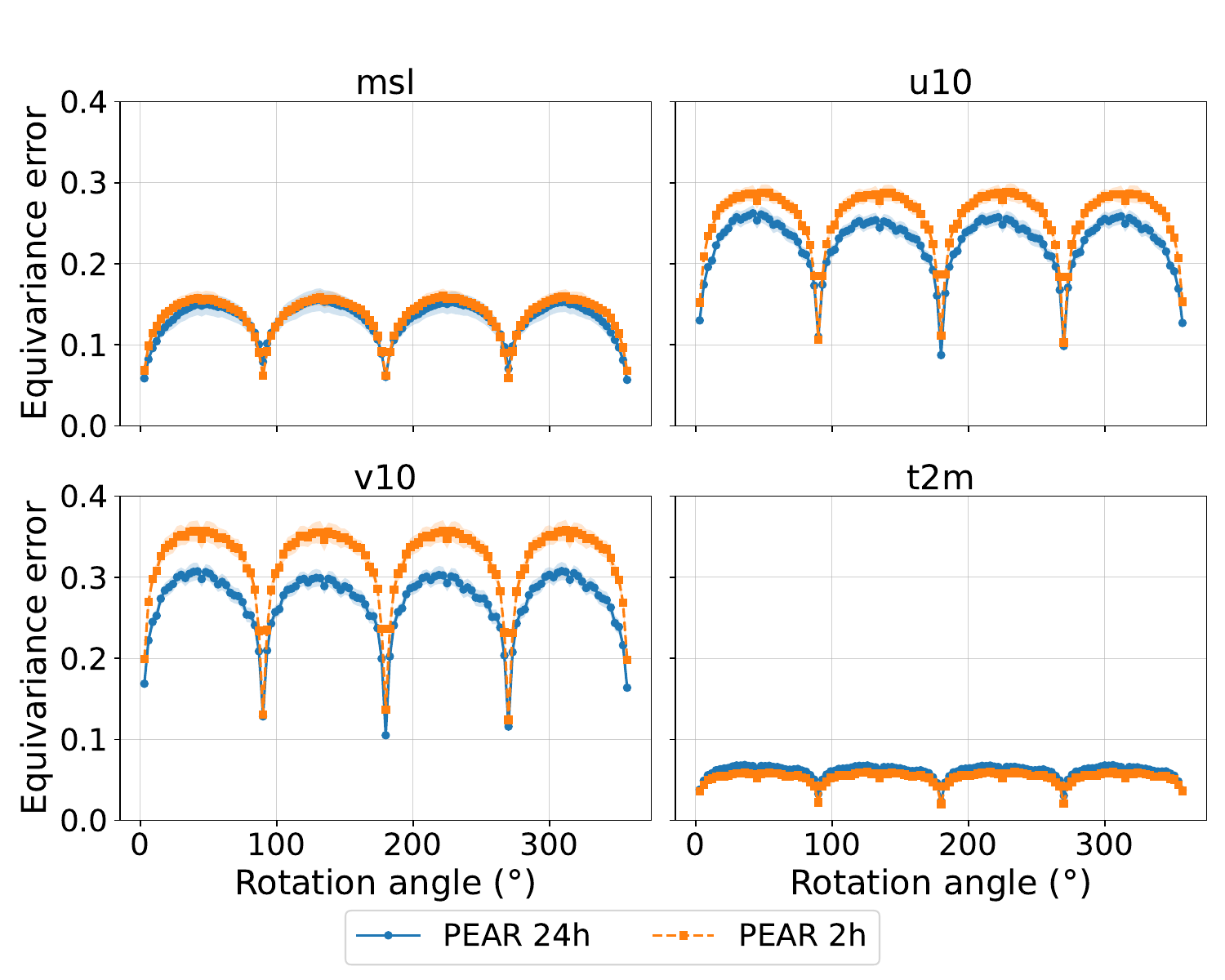}
    \caption{Mean equivariance error vs. rotation angle for surface  
  variables.}
    \label{fig:ee_surface}
\end{figure}
\begin{figure}
    \centering
    \includegraphics[height=0.95\textheight]{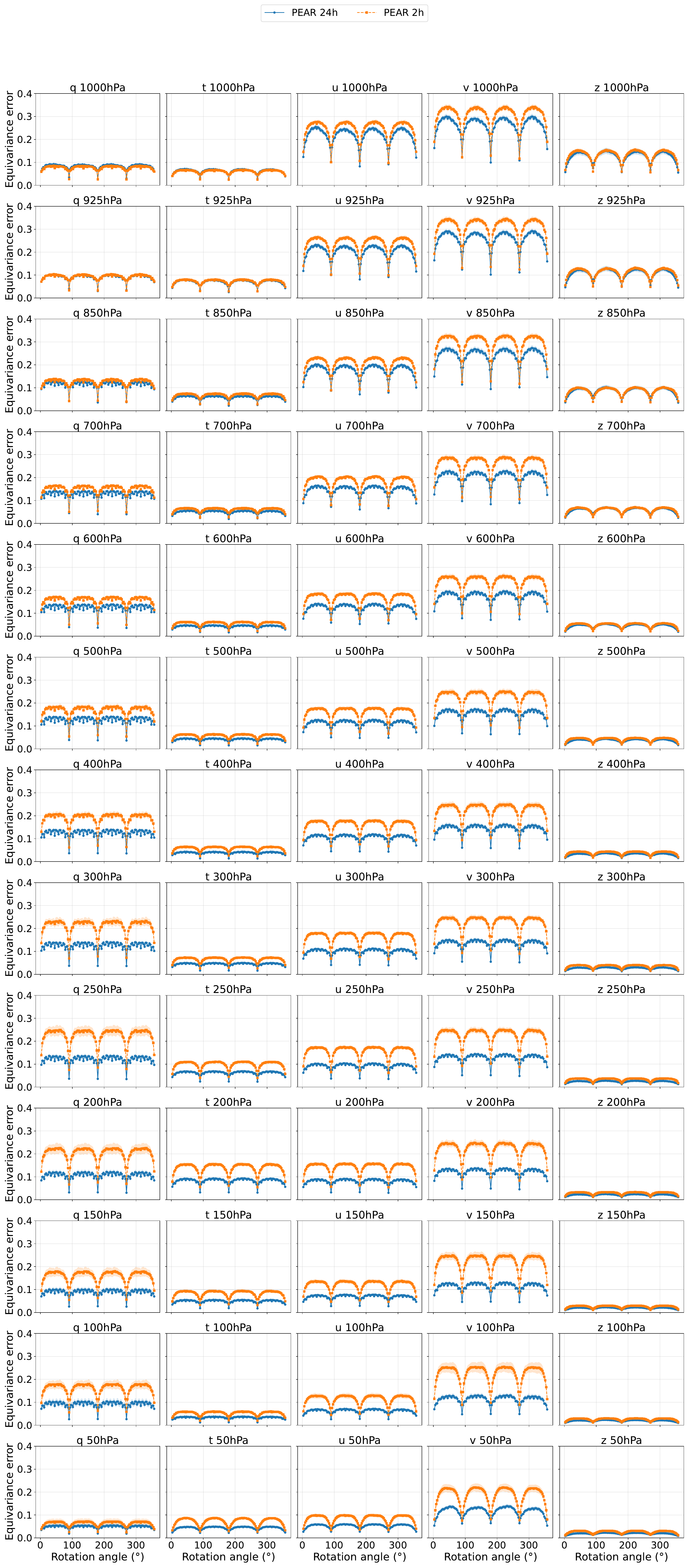}
    \caption{Mean equivariance error vs. rotation angle for 
  upper-level variables at each pressure level.}
    \label{fig:ee_upper}
\end{figure}

Figure \ref{fig:ee_surface} and Figure \ref{fig:ee_upper} show the per-channel equivariance error computed on PEAR after 200 training epochs, for the 2h time delta dataset and the 24h ERA5-Lite. The rotation angle $\theta$ is swept from $0^\circ$ to $360^\circ$ in steps of $3^\circ$, providing a fine-grained characterisation of the rotational symmetry in relation to the HEALPix discretisation of the sphere.
A common pattern emerges across all the prediction variables: the equivariance error peaks at $\theta \in \{45^\circ, 135^\circ, 225^\circ, 315^\circ\}$ and reaches its minima at $\theta \in \{0^\circ, 90^\circ, 180^\circ, 270^\circ, 360^\circ\}$. This eight-fold modulation is a direct signature of the underlying HEALPix grid, whose base resolution partitions the sphere into 12 diamond-shaped pixels arranged with a four-fold symmetry around the polar axis. Rotations by multiples of $90^\circ$ map the pixel centers to the pixel centers, so the rotated field aligns with the original sampling lattice and the error reduces to its irreducible component. In contrast, rotations by odd multiples of $45^\circ$ correspond to the maximal misalignment between the rotated and original grids, where each pixel center lies halfway between two cells of the reference partition.

A closer look at the results for the surface variables (Figure~\ref{fig:ee_surface}) reveals the absence of a common pattern across channels: each variable exhibits an independent behavior. In all panels, the blue curve corresponds to the model trained ot ERA5-Lite, while the orange curve refers to the new setup (2-hour sampling over 2012).
For \textit{msl} (mean sea-level pressure) and \textit{t2m} (2-metre temperature), the two setups yield nearly identical equivariance errors across all rotation angles, with the two curves essentially overlapping; the absolute error level differs between the two variables, as expected given their distinct physical nature. In contrast, \textit{u10} and \textit{v10} (wind speed along the surface) show a clear increase in equivariance error under the new sampling.
A possible interpretation is that the variables benefiting from the new setup are those whose evolution is more closely tied to the diurnal cycle, and therefore to the rotational symmetry that the 2-hour sampling is intended to exploit, while the wind components may be governed by factors that are less directly aligned with this symmetry. Under this hypothesis, the additional intra-day samples would not reinforce a rotational structure for \textit{u10} and \textit{v10}, but would instead expose the model to a broader range of configurations not related by rotations about the polar axis, increasing the apparent departure from $SO(2)$ equivariance. We present this only as a tentative explanation, as a rigorous assessment would require a dedicated analysis beyond the scope of this work.

\begin{figure}
    \centering
    \includegraphics[width=0.7\linewidth]{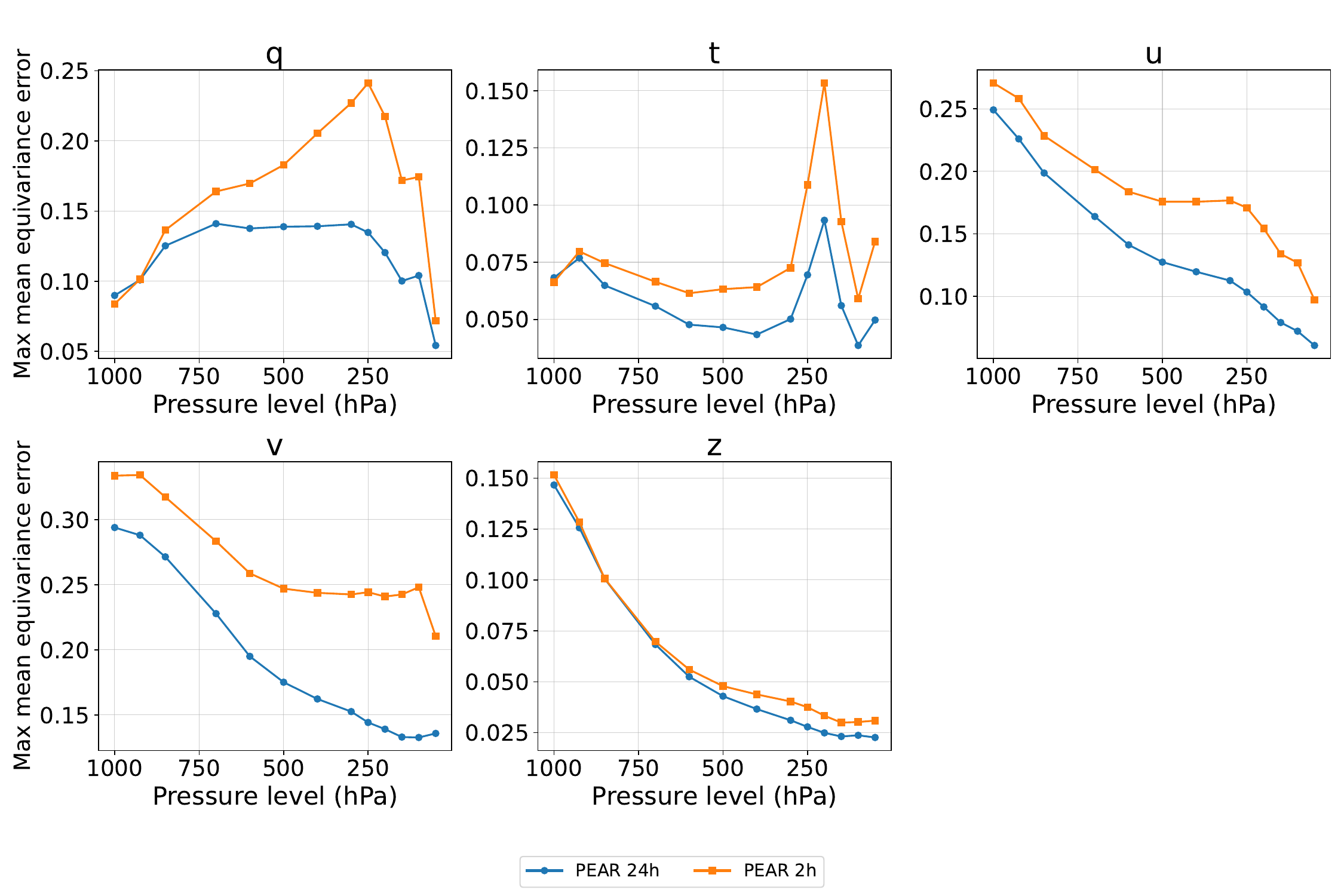}
    \caption{Peak mean equivariance error vs. pressure level for each
   upper-level variable. Each point is the maximum over      
  rotation angles of the epoch-mean equivariance error.}
    \label{fig:max_vs_level}
\end{figure}

Figure \ref{fig:max_vs_level} shows the peak mean equivariance error versus pressure level for each upper-level variable, where the peak is taken as the maximum over rotation angles of the epoch-mean equivariance error within each training
phase. Figure~\ref{fig:max_vs_level} shows that at the lowest level (1000~hPa), the variables \textit{q} (specific humidity), \textit{t} (temperature) and \textit{z} (geopotential) exhibit a similar equivariance error under the two setups, consistent with the behaviour observed for \textit{t2m} and \textit{msl} at the surface in Figure~\ref{fig:ee_surface}. As the altitude increases, however, the gap between the two setups widens progressively, with the original 24-hour sampling yielding a consistently lower equivariance error than the 2-hour setup. We hypothesize the training sets samples at a 2-hour interval are more closely correlated than the samples in ERA5-Lite and therefore provide a weaker training signal.

\begin{figure}
    \centering
    \includegraphics[width=0.7\linewidth]{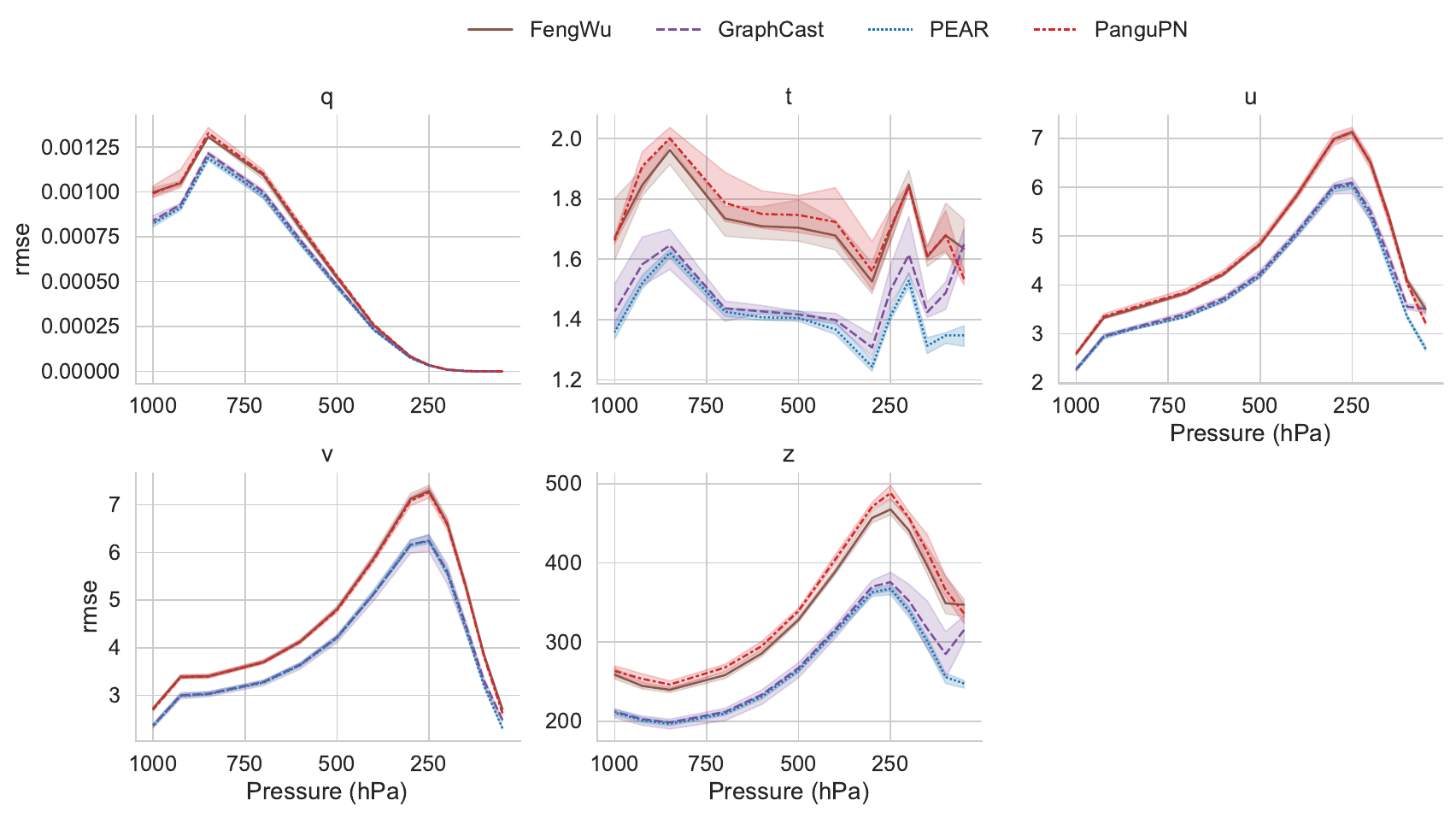}
    \caption{Average RMSE over different pressure layers, for the upper variables $q$, $t$, $u$, $v$, $z$, for different models}
    \label{fig:rmse_pressure}
\end{figure}

The dynamical variables \textit{u} and \textit{v} (wind speed component) follow the same trend already visible at the surface: the 24-hour setup yields a lower equivariance error at all levels, with the gap likewise increasing at higher altitudes.
A natural interpretation of these results is that the equivariance error decreases progressively with altitude due to lower influence of the asymmetric land mass. This trend is observed for \textit{u}, \textit{v} and \textit{z}, whose error decreases with height under both setups. However, the variables \textit{t} and \textit{q} do not follow this pattern, showing spikes at around the level 200~hPa. Comparing with overall model performance as measure by RMSE in Figure~\ref{fig:rmse_pressure}, we observe a similar spike in $t$ across all considered models, hinting at a common underlying reason. However, the RMSE profiles of $q$, \textit{u}, \textit{v} and \textit{z} do not seem to correlate well with the equivariance error results, indicating that the observed patterns are not caused by variances in overall model performance. This supports our hypothesis of the decrease land-mass effect for $u$, $v$ and $z$.

In conclusion, the comparison between the two sampling strategies reveals a more complex picture than initially expected. The 2-hour setup over a single year does not provide a uniform improvement in rotational equivariance: it does not show any clear advantage at either the surface or upper levels.
The main takeaway is that, for the subset of variables analyzed here, the original 24-hour sampling over 2007--2017 already provides a sufficiently expressive training signal: the model recovers the rotational structure of these channels well enough that the additional intra-day snapshots offer only a marginal improvement.%

\end{document}